\documentclass[runningheads]{llncs}

\usepackage{eccv}
\usepackage{eccvabbrv}

\usepackage{graphicx}
\usepackage{booktabs}
\usepackage[table]{xcolor}
\usepackage{arydshln}
\usepackage{multirow}
\usepackage{amssymb}

\usepackage{amsthm}
\usepackage{subcaption}
\usepackage{tabularx}
\newcolumntype{C}{>{\centering\arraybackslash}X}
\usepackage{pifont}
\usepackage{textcomp}
\usepackage{float}

\definecolor{darkgreen}{RGB}{0, 150, 0}
\definecolor{darkred}{RGB}{200, 0, 0}
\definecolor{darkorange}{RGB}{230, 115, 0}
\definecolor{darkyellow}{RGB}{180, 160, 0}

\renewcommand{\qedsymbol}{$\scriptstyle\square$}
\newcommand{\cmark}{\textcolor{darkgreen}{\ding{51}}}
\newcommand{\xmark}{\textcolor{darkred}{\ding{55}}}

\usepackage[accsupp]{axessibility}

\usepackage{hyperref}
\usepackage{orcidlink}

\makeatletter
\renewcommand{\contentsname}{Contents}
\newcommand{\supplementarytoc}{%
  \begingroup
  \def\authcount##1{}%
  \let\l@title\@gobbletwo
  \let\l@author\@gobbletwo
  \section*{\contentsname}
  \@starttoc{toc}%
  \endgroup
}
\makeatother

\begin{document}

% ===============================================================
% ARTÍCULO PRINCIPAL
% ===============================================================

% Desactiva las entradas del TOC durante el paper principal
\makeatletter
\let\ORIGaddcontentsline\addcontentsline
\renewcommand{\addcontentsline}[3]{}
\makeatother

\title{OPAL: Orthonormal Prototype Alignment Learning for Interpretable Image Classification}
\titlerunning{OPAL: Orthonormal Prototype Alignment Learning}

\author{Ilán Carretero\inst{1}\orcidlink{0009-0003-9474-6515} \and
Gustavo Jesús Angulo\inst{2}\orcidlink{0000-0001-6106-2692},\\
Rocío del Amor\inst{1,3}\orcidlink{0000-0002-5342-2093} \and
Valery Naranjo \inst{1,3}\orcidlink{0000-0002-0181-3412}}

\authorrunning{Carretero et al.}

\institute{CVBLab, HumanTech, Universitat Politècnica de
València, Valencia, Spain\\ \email{\{ilcarjuc,madeam2,vnaranjo\}@upv.es}\\ \and
CMA, Mines Paris, PSL University, Sophia Antipolis, France \\
\email{jesus.angulo\_lopez@minesparis.psl.eu} \and
Artikode Intelligence S.L., Valencia, Spain}

\maketitle

\begin{abstract}
  Prototypical part-based models provide explainable predictions by comparing input regions to learned prototypes. However, current approaches are burdened by complex, multi-stage training pipelines and heavily rely on auxiliary regularization to prevent prototype collapse. To overcome these limitations, we introduce Orthonormal Prototype Alignment Learning (OPAL), a single-stage, end-to-end framework that simplifies interpretable classification. Our approach anchors the latent space using predefined orthonormal bases, embedding each class within a dedicated subspace spanned by fixed part-prototypes. To achieve precise part localization, OPAL enforces spatial competition across feature maps. This mechanism isolates sparse, discriminative regions, directing each prototype to consistently attend to the same semantic concept across different images. By framing classification as a direct representation alignment task, our method eliminates the need for auxiliary losses. Extensive experiments on fine-grained benchmarks demonstrate that OPAL outperforms both its non-interpretable counterparts and state-of-the-art part-prototype methods, delivering granular visual explanations by explicitly revealing the specific image regions driving every prediction. Code is available at \url{https://github.com/ilancarretero/OPAL}.
\end{abstract}

\section{Introduction}
\label{sec:intro}

Deep neural networks have revolutionized the field of computer vision, delivering unprecedented performance across a spectrum of complex recognition tasks \cite{he2016deep, dosovitskiy2020image, liu2021swin, liu2022convnet}. Despite these advancements, the inherent opacity of these architectures presents a fundamental challenge \cite{longo2024explainable, kazmierczak2025explainability, choi2025icev2}. Beyond the critical trust issues arising in high-stakes deployment scenarios \cite{esteva2017dermatologist, codevilla2018end}, the inability to investigate the internal reasoning of a network limits the capacity to identify when and why the model fails, mitigate systematic errors, and potentially derive novel domain knowledge from learned feature representations. While post-hoc explanation methods, such as saliency maps \cite{ribeiro2016should, selvaraju2017grad}, attempt to interpret trained networks by highlighting relevant input regions, they have been shown to lack faithfulness to the actual decision-making process \cite{rudin2019stop, adebayo2018sanity}. Consequently, reliance on these approximations can be misleading, driving a paradigm shift towards inherently interpretable architectures where transparency is not a retrospective approximation, but a structural constraint enforced by design.

Among inherently interpretable models, Prototypical Part-based Networks (ProtoPNet) \cite{chen2019looks} and their recent variants \cite{rymarczyk2021protopshare, rymarczyk2022interpretable, nauta2023pip, pach2025lucidppn} have emerged as the dominant paradigm. These architectures ground their decision process in case-based reasoning, decomposing classification into a transparent matching process where local image regions are compared against a set of learned latent prototypes. However, despite their conceptual appeal, optimizing these frameworks involves a structural complexity significantly higher than their non-interpretable counterparts. As systematically illustrated in \cref{tab:intro_comparison}, state-of-the-art methods rely on cumbersome, multi-stage training pipelines that alternate between backbone optimization, prototype clustering, and discrete projection steps to map latent vectors to real training patches. This complexity stems fundamentally from the \textit{“moving target”} problem \cite{chen2019looks, wang2021interpretable}: since prototypes are learnable parameters residing in a continuously shifting latent space, the model struggles to simultaneously learn discriminative features and align prototypes to them. To mitigate semantic collapse, where prototypes converge to identical features \cite{rymarczyk2021protopshare}, existing approaches depend on multiple auxiliary regularization terms (e.g., separation, clustering, and diversity losses). This dependency significantly increases the complexity of hyperparameter tuning, as balancing these competing objectives requires meticulous dataset-specific calibration.

\begin{table}[tb]
\centering
\caption{High-level comparison between a standard CNN and representative inherently interpretable models, including our contribution, OPAL. \cmark\ indicates that the method satisfies the corresponding property, and \xmark\ otherwise. ``Aux-loss free'' denotes training without additional auxiliary regularizers beyond the standard classification objective. ``External-model free'' denotes no reliance on auxiliary pretrained models or external part-discovery modules.}
\label{tab:intro_comparison}
\renewcommand{\arraystretch}{1.2}
\scriptsize

\begin{tabularx}{\textwidth}{@{} l CCCC @{}}
\toprule
\multirow{2}{*}{\textbf{Method}} &
\textbf{Single-stage} &
\textbf{Aux-loss} &
\textbf{External-model} &
\textbf{Interpretable} \\
&
\textbf{training} &
\textbf{free} &
\textbf{free} &
\textbf{by design} \\
\midrule

Standard CNN
& \cmark & \cmark & \cmark & \xmark \\

\hdashline\noalign{\vskip 1mm}

ProtoPNet {\tiny\textcolor{gray}{NeurIPS '19}} \cite{chen2019looks}
& \xmark & \xmark & \cmark & \cmark \\

ProtoTree {\tiny\textcolor{gray}{CVPR '21}} \cite{nauta2021neural}
& \xmark & \cmark & \cmark & \cmark \\

ProtoPShare {\tiny\textcolor{gray}{KDD '21}} \cite{rymarczyk2021protopshare}
& \xmark & \xmark & \cmark & \cmark \\

TesNet {\tiny\textcolor{gray}{ICCV '21}} \cite{wang2021interpretable}
& \xmark & \xmark & \cmark & \cmark \\

ProtoPool {\tiny\textcolor{gray}{ECCV '22}} \cite{rymarczyk2022interpretable}
& \xmark & \xmark & \cmark & \cmark \\

PIP-Net {\tiny\textcolor{gray}{CVPR '23}} \cite{nauta2023pip}
& \xmark & \xmark & \cmark & \cmark \\

MCPNet {\tiny\textcolor{gray}{CVPR '24}} \cite{wang2024mcpnet}
& \cmark & \xmark & \cmark & \cmark \\

LucidPPN {\tiny\textcolor{gray}{ICLR '25}} \cite{pach2025lucidppn}
& \xmark & \xmark & \xmark & \cmark \\

\rowcolor{gray!15}
OPAL \textit{(Ours)}
& \cmark & \cmark & \cmark & \cmark \\
\bottomrule
\end{tabularx}
\end{table}

\begin{figure}[tb]
  \centering
  \includegraphics[width=\textwidth]{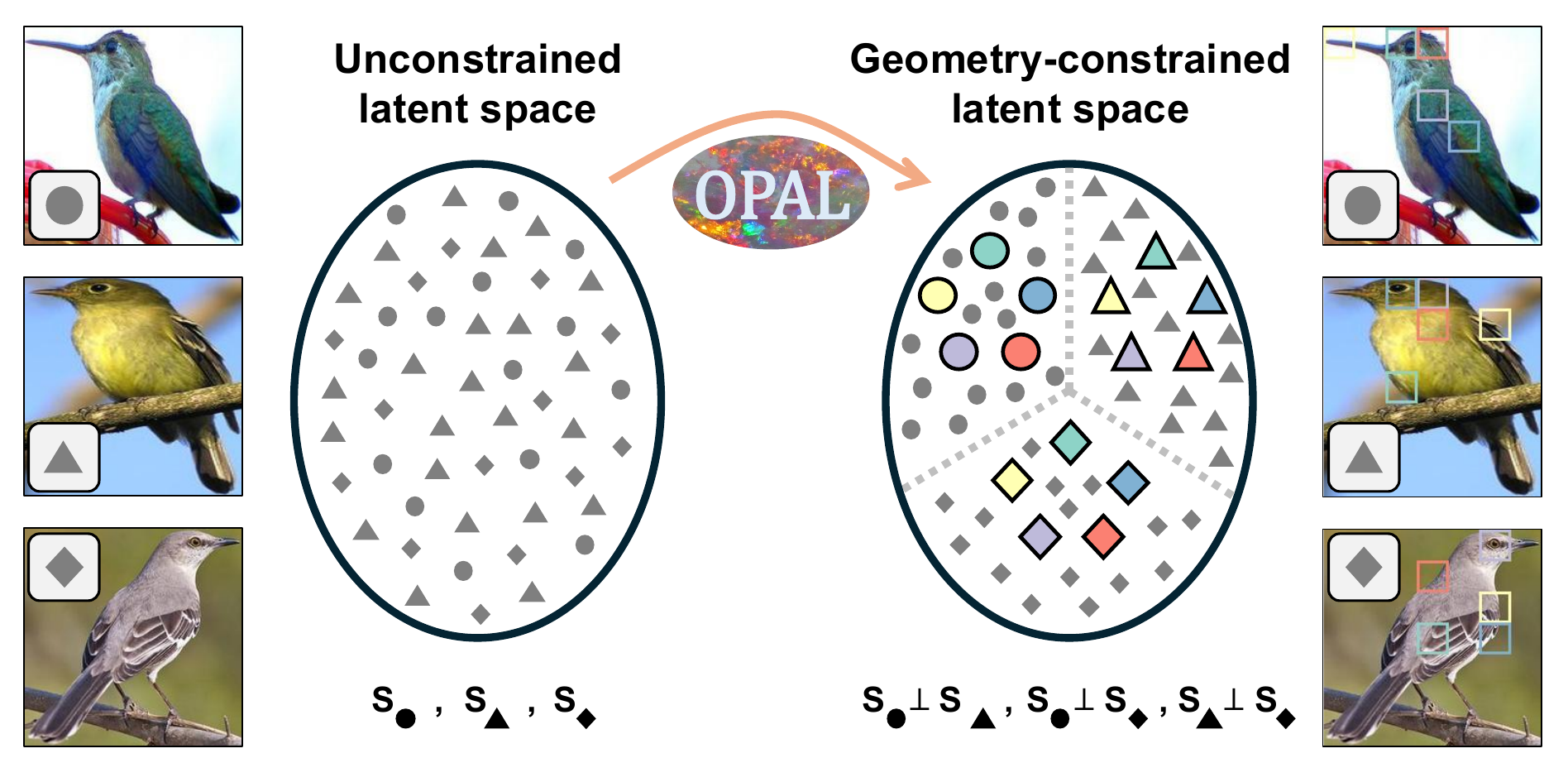}
  \caption{Conceptual overview of OPAL. Prior inherently interpretable approaches typically learn prototypes in an (\emph{unconstrained latent space}), where class evidence may overlap in a shared representation. OPAL instead imposes a (\emph{geometry-constrained latent space}) by assigning each class \(c\) to a dedicated orthonormal subspace \(\mathcal{S}_c \subset \mathbb{R}^{K}\), shown as \(\mathcal{S}_{\circ}, \mathcal{S}_{\triangle}, \mathcal{S}_{\diamond}\) in the figure. This structure encourages different prototype coordinates within \(\mathcal{S}_c\) to capture distinct semantic attributes and supports faithful part-based evidence, visualized as the image regions selected by the corresponding prototype activations.}
  \label{fig:opal_representation}
\end{figure}

In this work, we introduce a geometry-driven formulation for interpretable prototype-based classification that replaces unconstrained prototype discovery with structured subspace alignment. Leveraging the principle that fixed, maximally separable geometric constraints induce robust and transferable feature representations
\cite{mettes2019hyperspherical, papyan2020prevalence, hoffer2018fix, deng2019arcface}, we introduce Orthonormal Prototype Alignment Learning (OPAL). As illustrated in \cref{fig:opal_representation}, OPAL anchors each class to a pre-assigned orthonormal subspace and represents interpretability as alignment within this fixed geometry, rather than as the optimization of free prototype locations. This design decouples prototype geometry from representation learning and turns classification into a direct alignment task in a stable, theoretically grounded coordinate system.

To translate this geometric alignment into granular visual explanations, OPAL incorporates an intrinsic spatial competition mechanism. Unlike traditional approaches that rely on soft attention maps or auxiliary concentration losses to localize parts \cite{zheng2017learning,li2020deep}, our method enforces exclusiveness through channel-wise normalization followed by global feature selection. By compelling feature maps to compete for activation at each spatial location, OPAL prevents semantic concepts from overlapping. Consequently, a specific image region cannot simultaneously represent multiple distinct prototypes. This “winner-takes-all” dynamic, combined with the structural stability of the orthonormal basis, naturally drives the network to discover sparse and highly discriminative object parts without requiring part-level annotations or complex bounding-box supervision \cite{zhou2016learning, oquab2015object}.

The main contributions of this work are summarized as follows:
\begin{itemize}
    \item We introduce OPAL, a geometry-constrained approach to interpretable prototype based classification that moves from
unconstrained prototype discovery to structured alignment in a predefined orthonormal space. This design enables a
single-stage, end-to-end training pipeline without warm-up phases or alternating optimization steps.

    \item We identify that imposing strict orthogonality on the output space, combined with an explicit spatial competition
mechanism, is sufficient to induce semantic sparsity, eliminating the need for the auxiliary regularization terms (e.g., separation, clustering, or diversity losses) required by prior art to prevent prototype collapse. This results in a minimalist architecture that naturally isolates discriminative image regions by projecting them onto exclusive, preassigned semantic axes.

    \item We conduct an extensive evaluation across five fine-grained benchmarks, where OPAL attains leading performance among inherently interpretable models on most datasets. Moreover, OPAL consistently outperforms the corresponding non-interpretable CNN backbones across diverse architectures and model capacities. Qualitative results further indicate that prototype slots align with distinct semantic parts, yielding granular and conceptually meaningful visual explanations.
\end{itemize}

\section{Related work}
\label{sec:rel_work}

    \subsection{Interpretable Image Classification}
    \label{subec:interpretable_image_classification}

    The domain of interpretable image classification builds on the Prototypical Part Network (ProtoPNet) \cite{chen2019looks}, which introduced the case-based reasoning paradigm. However, its original formulation imposed a heavy optimization burden, relying on a multi-stage training pipeline and auxiliary regularization terms to align latent prototypes with training patches. Subsequent works aimed to mitigate prototype redundancy and improve reasoning efficiency: ProtoTree \cite{nauta2021neural} structured prototypes hierarchically, ProtoPShare \cite{rymarczyk2021protopshare} and ProtoPool \cite{rymarczyk2022interpretable} introduced mechanisms for prototype merging and differentiable assignment, and TesNet \cite{wang2021interpretable} represented each class through dedicated subspaces of a transparent embedding space. Despite their efficacy in compacting the latent space, these methods retain, and in cases like ProtoPool even increase, the complexity of the training objective by requiring extensive hyperparameter tuning for multiple competing auxiliary losses. 

    More recent state-of-the-art approaches prioritize the semantic grounding of prototypes to ensure they align with human-interpretable features. PIP-Net \cite{nauta2023pip} improves upon the visual coherence of explanations by framing classification as a sparse scoring task. By leveraging a self-supervised part-discovery objective, it ensures that prototypes correspond to consistent semantic patches, effectively identifying specific object parts rather than abstract latent vectors. However, achieving this alignment requires a decoupled two-stage training schedule and two auxiliary loss functions. MCPNet \cite{wang2024mcpnet} captures various semantic granularities through a multi-level hierarchy. However, its reliance on abstract concept prototypes rather than distinct prototypical parts compromises the interpretability of high-level attributes, making it difficult to differentiate the specific localized regions the model is attending to. Most recently, LucidPPN \cite{pach2025lucidppn} seeks to disentangle features such as color and shape, yet it crucially depends on an external part-discovery model, PDiscoNet \cite{van2023pdisconet}, for initial prototype alignment. This introduces an additional layer of auxiliary pre-training and parameter tuning. A separate line pursues prototype-based interpretability post-hoc, on top of frozen classifiers. Tan \etal \cite{tan2024post} decompose a trained classification head into part-prototypes using non-negative matrix factorization, while EPIC \cite{borycki2026epic} disentangles the final-layer channels with a learnable invertible transform, both producing prototype explanations without altering the original predictions. By operating on a frozen backbone, however, these methods explain a representation they cannot reshape, and the quality of their prototypes remains bounded by the features the model has already learned. In contrast, OPAL is interpretable by design while remaining single-stage. It removes the structural and procedural dependencies of prior ante-hoc methods and, unlike post-hoc approaches, jointly learns the representation and its explanation within a fixed, maximally separable class geometry. This matches the end-to-end simplicity of a standard CNN while providing robust, semantically meaningful interpretability.

    \subsection{Representation Learning and Metric Alignment}
    \label{subec:represention_learning}

    A complementary line of work studies how discriminative representations emerge from geometric constraints on the embedding space \cite{deng2019arcface, kim2020proxy}. Normalizing features to lie on the unit hypersphere and optimizing angular objectives has been highly effective in practice, particularly through margin-based formulations that operate directly in cosine space \cite{liu2017sphereface, wang2018cosface}. In parallel, contrastive learning provides a geometric view of representation quality through alignment and distributional spread on the hypersphere \cite{wang2020understanding}, and supervised contrastive objectives further strengthen class-level clustering in the normalized embedding space \cite{khosla2020supervised}. These perspectives motivate OPAL's use of a fixed, structured geometry where classification can be expressed as alignment in a normalized space.

    Metric learning methods also commonly replace standard linear classifiers with distance-based objectives defined over class representatives. Proxy-based approaches introduce class proxies to accelerate and stabilize metric learning, optimizing distances to representative points instead of relying on expensive sample mining \cite{movshovitz2017no, kim2020proxy, teh2020proxynca++}. Related observations further suggest that fixing the classifier geometry can be surprisingly effective, reducing the reliance on learning dense classification heads \cite{hoffer2018fix}. OPAL builds on these insights, but departs from prior proxy formulations by fixing the class geometry a priori through orthonormal subspaces and anchors, and by coupling this structure with competitive spatial evidence aggregation to obtain interpretable, part-based decisions under image-level supervision.

\section{Orthonormal Prototype Alignment Learning (OPAL)}
\label{sec:opal}

\begin{figure}[tb]
  \centering
  \includegraphics[width=\textwidth]{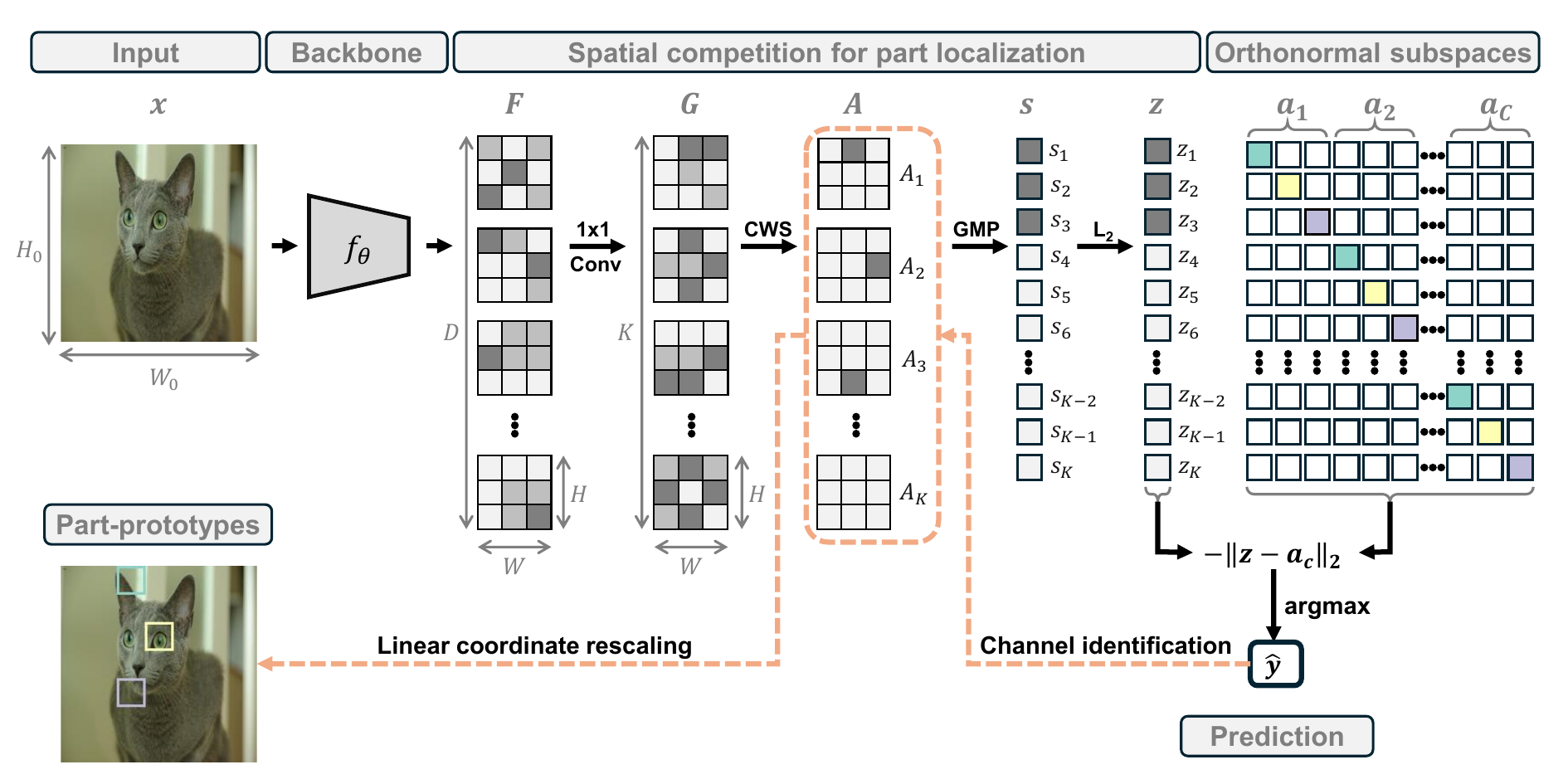}
  \caption{Overview of the OPAL framework. Given an input image \(x\in\mathbb{R}^{3\times H_0\times W_0}\), a convolutional backbone \(f_{\theta}\) produces a feature map \(F(x)\in\mathbb{R}^{D\times H\times W}\), which is projected via a \(1\times1\) convolution to \(G(x)\in\mathbb{R}^{K\times H\times W}\) with \(K=C\cdot m\) channels (\(m=3\) shown as a representative example). OPAL enforces part selection through (\emph{channel-wise softmax}, CWS), followed by (\emph{global max pooling}, GMP) and \(L_2\) normalization to obtain the embedding \(z\in\mathbb{R}^{K}\). Classification is performed by distance-based alignment to fixed class anchors \(\{a_c\}_{c=1}^{C}\) in (\emph{orthonormal subspaces}), yielding the prediction \(\hat{y}\). For interpretability, OPAL identifies the channels associated with \(\hat{y}\), and maps their spatial maximizers back to the input via (\emph{linear coordinate rescaling}) to produce part-prototype evidence.}
  \label{fig:opal_framework}
\end{figure}

An overview of OPAL is provided in \cref{fig:opal_framework}. The following subsections present the problem formulation and the key components of the method.

    \subsection{Preliminaries and Problem Formulation}
    \label{subsec:preliminaries_problem_formulation}

We address supervised fine-grained image classification under image-level supervision. Let
\(\mathcal{D}=\{(x_i,y_i)\}_{i=1}^{N}\) denote a dataset of \(N\) labeled images, where each image is \(x_i \in \mathbb{R}^{3\times H_0 \times W_0}\) with height \(H_0\) and width \(W_0\), and its label is \(y_i \in \{1,\dots,C\}\). The supervision is limited to the global class label. No additional annotations are assumed, including part labels, segmentation masks, or any other region-level information. Our objective is to learn a predictor that is accurate while remaining inherently interpretable by design. In addition to a predicted label \(\hat{y}\), the model must provide a transparent account of the evidence supporting its decision in terms of localized image regions.

We build on a convolutional feature extractor \(f_{\theta}\) that maps an input image \(x \in \mathbb{R}^{3\times H_0 \times W_0}\) to a final convolutional feature map \(F=f_{\theta}(x)\in \mathbb{R}^{D\times H \times W}\), where \(D\) is the number of channels and \(H\times W\) is the spatial resolution at the last convolutional stage. OPAL then refines this representation through a lightweight reparameterization and a competitive aggregation mechanism, and uses the resulting features to impose a structured output geometry that enables part-based
reasoning under image-level supervision.

    \subsection{Predefined Orthonormal Subspaces}
    \label{subsec:fixed_orthonormal_subspaces}

 OPAL fixes the geometry of the output space by assigning each class to a dedicated orthonormal subspace. Let \(m\in\mathbb{N}\) denote the number of prototype slots per class and define the structured output dimensionality as
\(K=C\cdot m\). The coordinate set \(\{1,\dots,K\}\) is arranged into \(C\) disjoint blocks \(\{B_c\}_{c=1}^{C}\), where \(B_c=\{(c-1)m+1,\dots,c\cdot m\}\) contains the \(m\) coordinates reserved for class \(c\). These blocks induce class-specific subspaces in \(\mathbb{R}^{K}\). Let \(\{e_k\}_{k=1}^{K}\) be the canonical basis of \(\mathbb{R}^{K}\). The subspace associated with class \(c\) is \(\mathcal{S}_c=\mathrm{span}\{e_k: k\in B_c\}\). By construction, \(\mathcal{S}_c \perp \mathcal{S}_{c'}\) for any \(c\neq c'\), and every vector \(z\in\mathbb{R}^{K}\) admits a unique decomposition into components lying in these mutually orthogonal subspaces.

Within each $\mathcal{S}_c$, we define $m$ fixed part-prototypes for class $c$, denoted by $\{p_{c,j}\}_{j=1}^{m}\subset\mathbb{R}^{K}$, which act as canonical prototype directions (one per slot). Concretely, the set \(\{p_{c,j}\}_{j=1}^{m}\) is given by \(p_{c,j}=e_{(c-1)m+j}\). These canonical directions are unit-norm and mutually orthogonal. Therefore, \(\{p_{c,j}\}\) forms an orthonormal family, yielding a maximally separated coordinate system for part-based evidence. For class-level decisions, each block is represented by a single class anchor \(a_c\in\mathbb{R}^{K}\), defined as the normalized block-uniform vector
\begin{equation}
a_c \;=\; \frac{1}{\sqrt{m}} \sum_{j=1}^m p_{c,j}.
\end{equation}
This choice satisfies \(\|a_c\|_2=1\) and \(\langle a_c,a_{c'}\rangle=0\) for \(c\neq c'\), anchoring each class to a unit vector and enforcing maximal angular separation between class representatives. The block-uniform anchor also
introduces an inductive bias that favors distributing evidence across the \(m\) slots of the correct class block rather than concentrating it on a single coordinate. A formal justification is provided in Sec.~A.1 of the supplementary material.

    \subsection{Spatial Competition for Part Localization}
    \label{subsec:spatial_competition}

OPAL promotes part-based evidence by enforcing explicit competition across channels at the last convolutional stage. Starting from the backbone feature map \(F\in\mathbb{R}^{D\times H\times W}\), a \(1\times1\) convolutional projection maps \(D\) channels to \(K=C\cdot m\) channels, aligned with the \(K\) prototype coordinates introduced in the previous subsection. This produces a projected activation tensor \(G\in\mathbb{R}^{K\times H\times W}\). The spatial
grid is indexed by \(\Omega=\{1,\dots,H\}\times\{1,\dots,W\}\), and \(G_k(u,v)\) denotes the activation of channel \(k\) at location \((u,v)\in\Omega\). Competition is enforced by applying a channel-wise softmax (CWS) over the \(K\) channels independently at every spatial location \((u,v)\). This yields normalized maps \(A\in[0,1]^{K\times H\times W}\) defined as:
\begin{equation}
A_k(u,v)=\frac{\exp(G_k(u,v))}{\sum_{j=1}^{K}\exp(G_j(u,v))},\qquad (u,v)\in\Omega.
\end{equation}
By construction, \(\sum_{k=1}^{K}A_k(u,v)=1\) for every \((u,v)\), coupling all channels through direct competition. As a result, a single spatial location cannot simultaneously support multiple channels, which encourages different channels to specialize in distinct discriminative regions.

A global representation is obtained via global max pooling (GMP) per channel. For each channel \(k\), OPAL aggregates spatial evidence and records the corresponding maximizer according to:
\begin{equation}
s_k=\max_{(u,v)\in\Omega}A_k(u,v),
\qquad
(u_k^\ast,v_k^\ast)=\arg\max_{(u,v)\in\Omega}A_k(u,v).
\end{equation}
Stacking \(\{s_k\}_{k=1}^{K}\) yields \(s\in\mathbb{R}^{K}\), and the final embedding is obtained by \(L_2\) normalization, \(z=s/\|s\|_2\in\mathbb{R}^{K}\). This design directly ties representation learning to localized evidence. Each coordinate \(z_k\) is sourced from a single spatial location \((u_k^\ast,v_k^\ast)\) selected by max pooling, and the corresponding image patch constitutes the exact evidence that determines the channel contribution used by the classifier.

    \subsection{Single-stage Optimization and Inference}
    \label{subsec:single_stage_opt}

OPAL is trained end-to-end in a single stage using a standard discriminative objective. The forward pass produces the normalized embedding \(z\in\mathbb{R}^{K}\) described in \cref{subsec:spatial_competition}. Classification is reformulated as metric alignment in the fixed output geometry defined by the orthonormal class anchors \(\{a_c\}_{c=1}^{C}\) introduced in \cref{subsec:fixed_orthonormal_subspaces}. Rather than learning a dense classifier,
OPAL directly uses distances to these anchors as predicted class-logit values. In particular, the logit for class \(c\) and the prediction are:
\begin{equation}
    \ell_c=-\|z-a_c\|_2,
    \qquad
    \hat{y} \;=\; \underset{c\in\{1,\dots,C\}}{\arg\max}\; \ell_c.
\end{equation}

Since both \(z\) and \(a_c\) are \(L_2\)-normalized, Euclidean distances depend only on the inter-vector angle. We analyze \(\|z-a_c\|_2^2\) since it induces the same ordering as \(\|z-a_c\|_2\) on \(\mathbb{R}_{\ge 0}\), and obtain \(\|z-a_c\|_2^2 = 2-2\langle z,a_c\rangle\). Thus, distance-based alignment is equivalent to maximizing cosine similarity on the unit hypersphere, with class separation enforced by anchor orthogonality. Parameter update is guided by the minimization of the cross-entropy (CE) with respect to distance-based logits, mathematically formulated as:
\begin{equation}
\mathcal{L}_{\mathrm{CE}}
=
-\log \frac{\exp(\ell_y)}{\sum_{c=1}^{C}\exp(\ell_c)},
\end{equation}
where \(\ell_y\) denotes the logit of the ground-truth class \(y\).
Gradients are backpropagated through the full model, while the anchors remain fixed. This leads to a single-stage pipeline without auxiliary clustering, projection steps, or additional regularization losses.

At inference time, OPAL produces visual explanations as a direct byproduct of the forward pass. The \(1\times1\) projection yields \(K=C\cdot m\) output channels that are in one-to-one correspondence with the fixed prototype coordinates defined in \cref{subsec:fixed_orthonormal_subspaces}. Since each class anchor \(a_c\) has support only on
its block \(B_c\), a prediction \(\hat{y}\) naturally selects the subset of channels \(k\in B_{\hat{y}}\) as the prototype slots used to form class evidence. For each selected channel \(k\), global max pooling identifies a unique maximizer \((u_k^\ast,v_k^\ast)\in\Omega\) that determines the pooled response \(s_k\), ensuring that the contribution of \(z_k\) is sourced from a single spatial location. Finally, the maximizer coordinates are mapped back to the input
resolution by linearly rescaling from the feature-map grid \(H\times W\) to \(H_0\times W_0\), yielding the localized evidence patches that constitute the explanation.

\section{Experiments}
\label{sec:experiments}

    \subsection{Experimental Setup}
    \label{subsec:preliminaries_problem_formulation}

        \subsubsection{Datasets.} We evaluate OPAL on five standard fine-grained benchmarks: CUB-200-2011 (200 bird species) \cite{wah2011caltech}, Stanford Cars (196 car models) \cite{krause20133d}, Oxford-IIIT Pet (37 breeds of cats and dogs) \cite{parkhi2012cats}, Stanford Dogs (120 dog breeds) \cite{khosla2011novel}, and Oxford Flowers-102 (102 flower categories) \cite{nilsback2008automated}. We follow the official train/test splits for all datasets. Additional dataset details and preprocessing steps are provided in Sec.~B.1 of the supplementary material.

        \subsubsection{Implementation Details.} We evaluate OPAL across three convolutional backbone families at three model scales each: ResNet-50/101/152 \cite{he2016deep}, EfficientNet-V2 S/M/L \cite{tan2021efficientnetv2}, and ConvNeXt Tiny/Small/Base \cite{liu2022convnet}. Images are resized to \(224\times224\) and augmented with TrivialAugment \cite{muller2021trivialaugment}, as in \cite{nauta2023pip}. We fix the number of prototypes per class to \(m=5\) across all datasets and backbones, with sensitivity to \(m\) analyzed in \cref{subsec:ablation studies}. Across all settings, models are trained for 30 epochs using Adam with a learning rate of \(1\times10^{-4}\) and a batch size of 64. All experiments were run on an NVIDIA A100 (40GB) GPU and repeated with three different random seeds, reporting results averaged across runs. In addition to mean performance, we report the absolute percentage-point difference relative to the corresponding CNN backbone trained under the same protocol. Additional implementation details are provided in Sec.~B.2 of the supplementary material.

    \subsection{Quantitative Evaluation}
    \label{subsec:quantitative_evaluation}

 \subsubsection{Comparison with Inherently Interpretable Models.}

    \cref{tab:sota_comparison} reports top-1 accuracy on five fine-grained benchmarks and compares OPAL against eight interpretable methods, together with a non-interpretable counterpart. We use ConvNeXt-Tiny as the baseline backbone, consistent with recent prototype-based approaches that report results on this architecture \cite{nauta2023pip, wang2024mcpnet, pach2025lucidppn}, enabling direct backbone-aligned comparisons. Overall, this setting provides a broad side-by-side evaluation across datasets and representative interpretable baselines.

    OPAL achieves the best performance on 4 out of 5 datasets and ranks third on CARS, behind TesNet\cite{wang2021interpretable} and LucidPPN\cite{pach2025lucidppn}. Notably, OPAL is the only method that consistently surpasses the corresponding CNN backbone across all benchmarks, with an average improvement of +1.56 absolute percentage points over the baseline. These results suggest that imposing a fixed, highly discriminative latent space enables a single-stage training pipeline without auxiliary losses, thereby improving not only interpretability but also performance.

     \begin{table}[tb]
\centering
\caption{Top-1 accuracy (\%) comparison on five fine-grained benchmarks between OPAL and representative inherently interpretable models. \textcolor{darkgreen}{$\uparrow$} and \textcolor{darkred}{$\downarrow$} denote the absolute percentage point difference with respect to the ConvNeXt-Tiny baseline. ProtoTree exhibits low performance on DOGS and FLOWER, a behavior also observed on these datasets in \cite{pach2025lucidppn} and reported on another fine-grained benchmark in \cite{wang2024mcpnet}.}
\label{tab:sota_comparison}
\renewcommand{\arraystretch}{1.2}
\scriptsize 

\begin{tabularx}{\textwidth}{@{} l CCCCC @{}}
\toprule
\multirow{2}{*}{\textbf{Method}} & \multicolumn{5}{c}{\textbf{Accuracy ($\uparrow$)}} \\
\cmidrule(l){2-6}
& \textbf{CUB} & \textbf{CARS} & \textbf{PETS} & \textbf{DOGS} & \textbf{FLOWER} \\
\midrule

% Baseline
Baseline (ConvNext-Tiny)
& 84.2 
& 87.8 
& 91.0 
& 85.3 
& 95.6 \\

\hdashline\noalign{\vskip 1mm}

% SOTA Models
ProtoPNet {\tiny\textcolor{gray}{NeurIPS '19}} \cite{chen2019looks} 
& 79.2{\tiny\color{darkred}{$\downarrow$5.0}} 
& 86.1{\tiny\color{darkred}{$\downarrow$1.7}} 
& 80.9{\tiny\color{darkred}{$\downarrow$10.1}} 
& 77.4{\tiny\color{darkred}{$\downarrow$7.9}} 
& 92.1{\tiny\color{darkred}{$\downarrow$3.5}} \\

ProtoTree {\tiny\textcolor{gray}{CVPR '21}} \cite{nauta2021neural} 
& 82.2{\tiny\color{darkred}{$\downarrow$2.0}} 
& 86.6{\tiny\color{darkred}{$\downarrow$1.2}} 
& 75.0{\tiny\color{darkred}{$\downarrow$16.0}} 
& 58.9{\tiny\color{darkred}{$\downarrow$26.4}} 
& 17.5{\tiny\color{darkred}{$\downarrow$78.1}} \\

ProtoPShare {\tiny\textcolor{gray}{KDD '21}} \cite{rymarczyk2021protopshare} 
& 74.7{\tiny\color{darkred}{$\downarrow$9.5}} 
& 86.4{\tiny\color{darkred}{$\downarrow$1.4}} 
& 81.2{\tiny\color{darkred}{$\downarrow$9.8}}
& 74.1{\tiny\color{darkred}{$\downarrow$11.2}} 
& 90.3{\tiny\color{darkred}{$\downarrow$5.3}} \\

TesNet {\tiny\textcolor{gray}{ICCV '21}} \cite{wang2021interpretable} 
& 84.6{\tiny\color{darkgreen}{$\uparrow$0.4}} 
& \textbf{92.6}{\tiny\color{darkgreen}{$\uparrow$4.8}} 
& 88.7{\tiny\color{darkred}{$\downarrow$2.3}}
& 80.7{\tiny\color{darkred}{$\downarrow$4.6}} 
& 83.4{\tiny\color{darkred}{$\downarrow$12.2}} \\

ProtoPool {\tiny\textcolor{gray}{ECCV '22}} \cite{rymarczyk2022interpretable} 
& 85.5{\tiny\color{darkgreen}{$\uparrow$1.3}} 
& 88.9{\tiny\color{darkgreen}{$\uparrow$1.1}} 
& 80.8{\tiny\color{darkred}{$\downarrow$10.2}} 
& 71.7{\tiny\color{darkred}{$\downarrow$13.6}} 
& 92.7{\tiny\color{darkred}{$\downarrow$2.9}} \\

PIP-Net {\tiny\textcolor{gray}{CVPR '23}} \cite{nauta2023pip} 
& 84.3{\tiny\color{darkgreen}{$\uparrow$0.1}} 
& 88.2{\tiny\color{darkgreen}{$\uparrow$0.4}} 
& 92.0{\tiny\color{darkgreen}{$\uparrow$1.0}} 
& 80.8{\tiny\color{darkred}{$\downarrow$4.5}} 
& 91.8{\tiny\color{darkred}{$\downarrow$3.8}} \\

MCPNet {\tiny\textcolor{gray}{CVPR '24}} \cite{wang2024mcpnet} 
& 83.5{\tiny\color{darkred}{$\downarrow$0.7}} 
& 83.2{\tiny\color{darkred}{$\downarrow$4.6}} 
& 91.3{\tiny\color{darkgreen}{$\uparrow$0.3}} 
& 76.5{\tiny\color{darkred}{$\downarrow$8.8}} 
& 94.7{\tiny\color{darkred}{$\downarrow$0.9}} \\

LucidPPN {\tiny\textcolor{gray}{ICLR '25}} \cite{pach2025lucidppn} 
& 81.5{\tiny\color{darkred}{$\downarrow$2.7}} 
& 91.6 {\tiny\color{darkgreen}{$\uparrow$3.8}} 
& 92.1{\tiny\color{darkgreen}{$\uparrow$1.1}} 
& 79.5{\tiny\color{darkred}{$\downarrow$5.8}} 
& 95.0{\tiny\color{darkred}{$\downarrow$0.6}} \\

% Ours
\rowcolor{gray!15}
OPAL \textit{(Ours)} 
& \textbf{85.6}{\tiny\color{darkgreen}{$\uparrow$1.4}} 
& 90.3{\tiny\color{darkgreen}{$\uparrow$2.5}} 
& \textbf{92.8}{\tiny\color{darkgreen}{$\uparrow$1.8}} 
& \textbf{86.9}{\tiny\color{darkgreen}{$\uparrow$1.6}} 
& \textbf{96.1}{\tiny\color{darkgreen}{$\uparrow$0.5}} \\
\bottomrule
\end{tabularx}
\end{table}

        \subsubsection{Backbone Generalization and Scalability.}

\cref{tab:backbone_scalability} evaluates OPAL as a plug-in module across three convolutional backbone families and three model scales per family, always comparing against the corresponding CNN backbone trained under the same protocol. This setting isolates the effect of OPAL from backbone-specific design choices and tests whether the approach transfers across architectural families and capacities.

Overall, OPAL improves or matches the non-interpretable counterpart in 39 out of 45 backbone–dataset experiments, with an average gain of +2.94 absolute percentage points in top-1 accuracy across all datasets and backbones. In the few cases where OPAL does not outperform the baseline, the largest drop is 1.0 percentage points, indicating that the method is broadly robust and does not trade performance for interpretability when integrated into diverse convolutional architectures.

\begin{table}[tb]
\centering
\caption{Backbone generalization and scalability of OPAL across three convolutional backbone families and three model scales per family. Each backbone is evaluated in its standard form and with OPAL integrated (shaded rows). \textcolor{darkgreen}{$\uparrow$} and \textcolor{darkred}{$\downarrow$} denote the absolute percentage-point difference with respect to the corresponding CNN backbone.}
\label{tab:backbone_scalability}
\renewcommand{\arraystretch}{1.2}
\scriptsize 

\begin{tabularx}{\textwidth}{@{} l CCCCC @{}}
\toprule
\multirow{2}{*}{\textbf{Backbone}} & \multicolumn{5}{c}{\textbf{Accuracy ($\uparrow$)}} \\
\cmidrule(l){2-6}
& \textbf{CUB} & \textbf{CARS} & \textbf{PETS} & \textbf{DOGS} & \textbf{FLOWER} \\
\midrule

% ResNet-50
ResNet-50 
& 72.6 & 71.9 & 89.5 & 81.3 & 92.8 \\
\rowcolor{gray!15}
\multicolumn{1}{c}{\cellcolor{gray!15}w/ OPAL} 
& 80.8{\tiny\color{darkgreen}{$\uparrow$8.2}} 
& 84.3{\tiny\color{darkgreen}{$\uparrow$12.4}} 
& 89.8{\tiny\color{darkgreen}{$\uparrow$0.3}} 
& 81.0{\tiny\color{darkred}{$\downarrow$0.3}} 
& 92.7{\tiny\color{darkred}{$\downarrow$0.1}} \\

\hdashline\noalign{\vskip 1mm}

% ResNet-101
ResNet-101 
& 74.9 & 73.8 & 88.5 & 83.5 & 91.6 \\
\rowcolor{gray!15}
\multicolumn{1}{c}{\cellcolor{gray!15}w/ OPAL} 
& 81.4{\tiny\color{darkgreen}{$\uparrow$6.5}} 
& 83.9{\tiny\color{darkgreen}{$\uparrow$10.1}} 
& 91.0{\tiny\color{darkgreen}{$\uparrow$2.5}} 
& 83.0{\tiny\color{darkred}{$\downarrow$0.5}} 
& 93.0{\tiny\color{darkgreen}{$\uparrow$1.4}} \\

\hdashline\noalign{\vskip 1mm}

% ResNet-152
ResNet-152 
& 76.0 & 73.7 & 90.3 & 83.1 & 92.4 \\
\rowcolor{gray!15}
\multicolumn{1}{c}{\cellcolor{gray!15}w/ OPAL} 
& 80.7{\tiny\color{darkgreen}{$\uparrow$4.7}} 
& 85.1{\tiny\color{darkgreen}{$\uparrow$11.4}} 
& 90.1{\tiny\color{darkred}{$\downarrow$0.2}} 
& 84.1{\tiny\color{darkgreen}{$\uparrow$1.0}} 
& 93.8{\tiny\color{darkgreen}{$\uparrow$1.4}} \\

\hdashline\noalign{\vskip 1mm}

% EfficientNet-V2 S
EfficientNet-V2 S 
& 78.3 & 76.7 & 89.6 & 84.5 & 95.3 \\
\rowcolor{gray!15}
\multicolumn{1}{c}{\cellcolor{gray!15}w/ OPAL} 
& 83.5{\tiny\color{darkgreen}{$\uparrow$5.2}} 
& 87.2{\tiny\color{darkgreen}{$\uparrow$10.5}} 
& 90.4{\tiny\color{darkgreen}{$\uparrow$0.8}} 
& 85.8{\tiny\color{darkgreen}{$\uparrow$1.3}} 
& 95.3{\tiny\color{gray}{- 0.0}} \\

\hdashline\noalign{\vskip 1mm}

% EfficientNet-V2 M
EfficientNet-V2 M 
& 79.4 & 80.3 & 89.0 & 83.3 & 93.9 \\
\rowcolor{gray!15}
\multicolumn{1}{c}{\cellcolor{gray!15}w/ OPAL} 
& 82.6{\tiny\color{darkgreen}{$\uparrow$3.2}} 
& 82.7{\tiny\color{darkgreen}{$\uparrow$2.4}} 
& 90.6{\tiny\color{darkgreen}{$\uparrow$1.6}} 
& 85.7{\tiny\color{darkgreen}{$\uparrow$2.4}} 
& 92.9{\tiny\color{darkred}{$\downarrow$1.0}} \\

\hdashline\noalign{\vskip 1mm}

% EfficientNet-V2 L
EfficientNet-V2 L 
& 84.1 & 85.9 & 90.0 & 81.4 & 94.9 \\
\rowcolor{gray!15}
\multicolumn{1}{c}{\cellcolor{gray!15}w/ OPAL} 
& 86.9{\tiny\color{darkgreen}{$\uparrow$2.8}} 
& 89.4{\tiny\color{darkgreen}{$\uparrow$3.5}} 
& 90.0{\tiny\color{gray}{- 0.0}} 
& 83.1{\tiny\color{darkgreen}{$\uparrow$1.7}} 
& 96.6{\tiny\color{darkgreen}{$\uparrow$1.7}} \\

\hdashline\noalign{\vskip 1mm}

% ConvNext-Tiny
ConvNeXt-Tiny 
& 84.2 & 87.8 & 91.0 & 85.3 & 95.6 \\
\rowcolor{gray!15}
\multicolumn{1}{c}{\cellcolor{gray!15}w/ OPAL} 
& 85.6{\tiny\color{darkgreen}{$\uparrow$1.4}} 
& 90.3{\tiny\color{darkgreen}{$\uparrow$2.5}} 
& 92.8{\tiny\color{darkgreen}{$\uparrow$1.8}} 
& 86.9{\tiny\color{darkgreen}{$\uparrow$1.6}} 
& 96.1{\tiny\color{darkgreen}{$\uparrow$0.5}} \\

\hdashline\noalign{\vskip 1mm}

% ConvNext-Small
ConvNeXt-Small 
& 81.7 & 83.9 & 91.8 & 86.5 & 95.3 \\
\rowcolor{gray!15}
\multicolumn{1}{c}{\cellcolor{gray!15}w/ OPAL} 
& 85.6{\tiny\color{darkgreen}{$\uparrow$3.9}} 
& 90.6{\tiny\color{darkgreen}{$\uparrow$6.7}} 
& 93.4{\tiny\color{darkgreen}{$\uparrow$1.6}} 
& 87.7{\tiny\color{darkgreen}{$\uparrow$1.2}} 
& 95.5{\tiny\color{darkgreen}{$\uparrow$0.2}} \\

\hdashline\noalign{\vskip 1mm}

% ConvNext-Base
ConvNeXt-Base 
& 82.1 & 82.1 & 92.0 & 88.9 & 97.2 \\
\rowcolor{gray!15}
\multicolumn{1}{c}{\cellcolor{gray!15}w/ OPAL} 
& 86.3{\tiny\color{darkgreen}{$\uparrow$4.2}} 
& 91.4{\tiny\color{darkgreen}{$\uparrow$9.3}} 
& 93.9{\tiny\color{darkgreen}{$\uparrow$1.9}} 
& 90.2{\tiny\color{darkgreen}{$\uparrow$1.3}} 
& 96.6{\tiny\color{darkred}{$\downarrow$0.6}} \\

\bottomrule
\end{tabularx}
\end{table}

 \begin{figure}[tb]
  \centering
  \includegraphics[width=\textwidth]{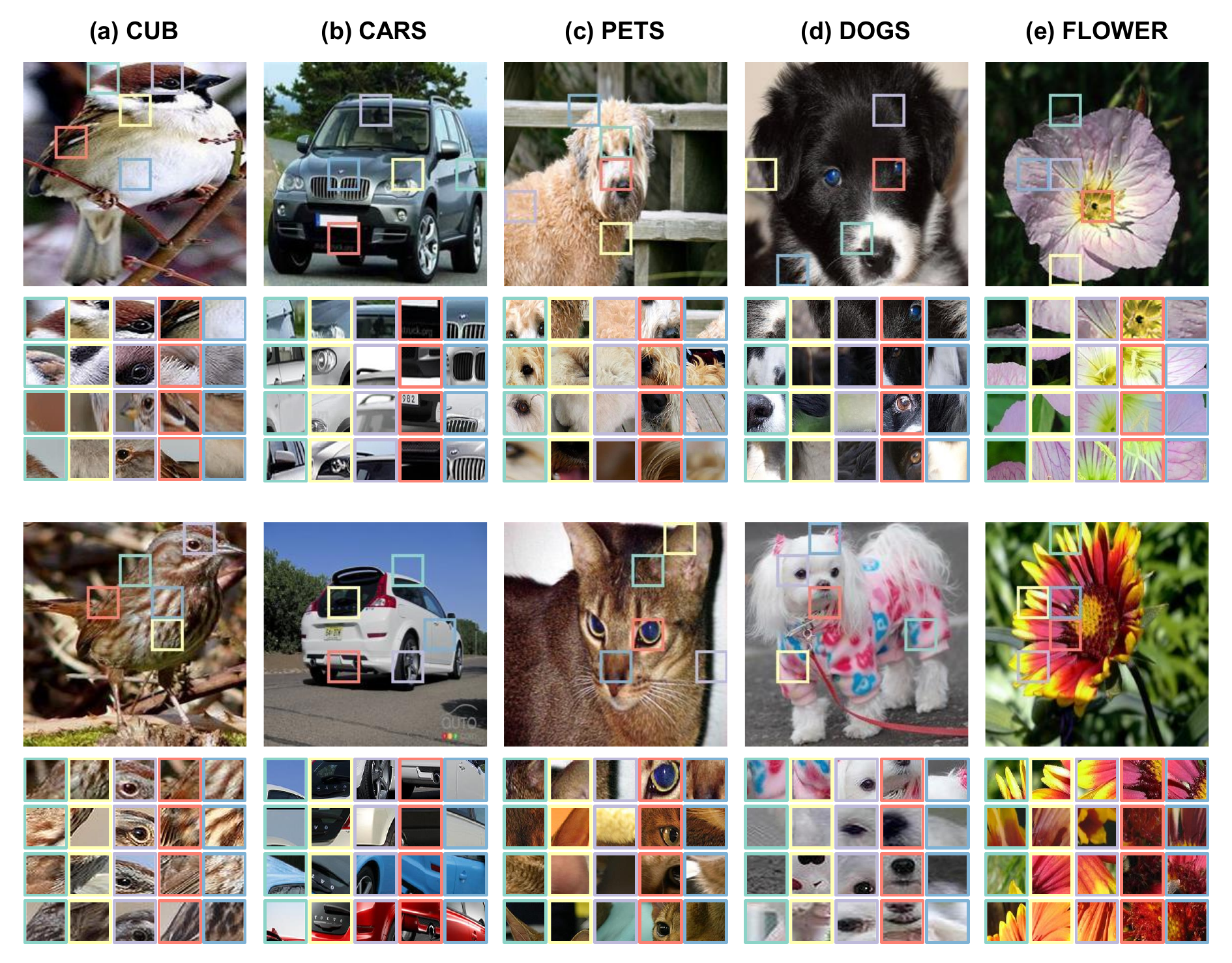}
  \caption{Qualitative prototype-based explanations produced by OPAL on five fine-grained benchmarks (\(m=5\)). For each dataset, we show two test images and the corresponding prototype-wise evidence. Colored boxes indicate the spatial maximizers selected for each prototype. The patch grids visualize the evidence associated with each prototype, with the first row extracted from the shown image and subsequent rows taken from other images of the same predicted
  class.}
\label{fig:qualitative_prototypes}
\end{figure}

    \subsection{Qualitative Interpretability Analysis}
    \label{subsec:qualitative_evaluation}

Interpretable predictions are essential to understand which visual evidence drives a model decision. \cref{fig:qualitative_prototypes} illustrates OPAL explanations across all five datasets. Across benchmarks, OPAL yields non-collapsed prototypes that activate on multiple, distinct regions of the object within a single image. In each example, the \(m=5\) prototypes provide a compact set of complementary, localized evidence spanning different discriminative parts of the object. This is especially important in fine-grained recognition, where class separation is often driven by subtle part-level attributes rather than a single salient region.

Beyond diversity, the retrieved evidence is also consistent within class across images, suggesting that individual prototypes acquire stable semantic roles. The same prototype repeatedly activates on visually and semantically related parts for different instances, indicating that the fixed class-wise orthonormal structure encourages different prototype directions to align with recurring attributes (e.g., ears for Abyssinian cats in PETS or eyes for Maltese dogs in DOGS). Further qualitative results are provided in the supplementary material, including additional prototype-based explanations (Sec.~C.1), the qualitative effect of the number of prototypes per class (Sec.~C.2), and an analysis of prototype collapse (Sec.~D.1). As a complementary quantitative assessment, and since part-level annotations are available only for CUB-200-2011, we report prototype consistency and stability on this dataset in Sec.~E.3 of the supplementary material, where OPAL attains the second-best results among prototype-based methods.

    \subsection{Ablation Studies}
    \label{subsec:ablation studies}

\begin{table}[tb]
\centering
\caption{Architectural ablation of OPAL components on five fine-grained benchmarks, reported as top-1 accuracy (\%). Columns correspond to the \(1\times1\) convolutional projection (Proj.), global max pooling (GMP), channel-wise softmax (CWS), and orthonormal anchors (Orth.). \cmark\ and \xmark\ indicate whether a component is enabled or not. When GMP is disabled, global average pooling is used. The all-crosses configuration (\xmark\xmark\xmark\xmark) corresponds to the convolutional baseline, while the all-checks configuration (\cmark\cmark\cmark\cmark) corresponds to full OPAL. \textcolor{darkgreen}{$\uparrow$} and \textcolor{darkred}{$\downarrow$} denote absolute percentage-point differences with respect to the baseline.}
\label{tab:ablation_full}
\renewcommand{\arraystretch}{1.2}
\scriptsize 

\begin{tabularx}{\textwidth}{@{} cccc CCCCC @{}}
\toprule
\multicolumn{4}{c}{\textbf{Architectural Components}} & \multicolumn{5}{c}{\textbf{Accuracy ($\uparrow$)}} \\
\cmidrule(r){1-4} \cmidrule(l){5-9}
\textbf{Proj.} & \textbf{GMP} & \textbf{CWS} & \textbf{Orth.} 
& \textbf{CUB} & \textbf{CARS} & \textbf{PETS} & \textbf{DOGS} & \textbf{FLOWER} \\
\midrule

% Row 1: Baseline (No, No, No, No)
\xmark & \xmark & \xmark & \xmark 
& 84.2 & 87.8 & 91.0 & 85.3 & 95.6 \\

\hdashline\noalign{\vskip 1mm}

% Row 2: Yes, No, No, No
\cmark & \xmark & \xmark & \xmark 
& 82.3{\tiny\color{darkred}{$\downarrow$1.9}} 
& 86.3{\tiny\color{darkred}{$\downarrow$1.5}} 
& 92.0{\tiny\color{darkgreen}{$\uparrow$1.0}} 
& 86.2{\tiny\color{darkgreen}{$\uparrow$0.9}} 
& 96.3{\tiny\color{darkgreen}{$\uparrow$0.7}} \\

% Row 3: No, Yes, No, No
\xmark & \cmark & \xmark & \xmark 
& 79.6{\tiny\color{darkred}{$\downarrow$4.6}} 
& 75.7{\tiny\color{darkred}{$\downarrow$12.1}} 
& 88.9{\tiny\color{darkred}{$\downarrow$2.1}} 
& 84.1{\tiny\color{darkred}{$\downarrow$1.2}} 
& 93.4{\tiny\color{darkred}{$\downarrow$2.2}} \\

% Row 4: No, No, Yes, No
\xmark & \xmark & \cmark & \xmark 
& 57.4{\tiny\color{darkred}{$\downarrow$26.8}} 
& 48.9{\tiny\color{darkred}{$\downarrow$38.9}} 
& 88.1{\tiny\color{darkred}{$\downarrow$2.9}} 
& 74.6{\tiny\color{darkred}{$\downarrow$10.7}} 
& 91.5{\tiny\color{darkred}{$\downarrow$4.1}} \\

% Row 5: Yes, Yes, No, No
\cmark & \cmark & \xmark & \xmark 
& 77.1{\tiny\color{darkred}{$\downarrow$7.1}} 
& 73.5{\tiny\color{darkred}{$\downarrow$14.3}} 
& 91.6{\tiny\color{darkgreen}{$\uparrow$0.6}} 
& 86.5{\tiny\color{darkgreen}{$\uparrow$1.2}} 
& 94.2{\tiny\color{darkred}{$\downarrow$1.4}} \\

% Row 6: No, Yes, Yes, No
\xmark & \cmark & \cmark & \xmark 
& 54.6{\tiny\color{darkred}{$\downarrow$29.6}} 
& 31.9{\tiny\color{darkred}{$\downarrow$55.9}} 
& 86.2{\tiny\color{darkred}{$\downarrow$4.8}} 
& 80.2{\tiny\color{darkred}{$\downarrow$5.1}} 
& 86.3{\tiny\color{darkred}{$\downarrow$9.3}} \\

% Row 7: Yes, No, Yes, No
\cmark & \xmark & \cmark & \xmark 
& 73.7{\tiny\color{darkred}{$\downarrow$10.5}} 
& 70.5{\tiny\color{darkred}{$\downarrow$17.3}} 
& 91.1{\tiny\color{darkgreen}{$\uparrow$0.1}} 
& 84.7{\tiny\color{darkred}{$\downarrow$0.6}} 
& 95.0{\tiny\color{darkred}{$\downarrow$0.6}} \\

% Row 8: Yes, No, No, Yes
\cmark & \xmark & \xmark & \cmark 
& 85.0{\tiny\color{darkgreen}{$\uparrow$0.8}} 
& 90.6{\tiny\color{darkgreen}{$\uparrow$2.8}} 
& 92.6{\tiny\color{darkgreen}{$\uparrow$1.6}} 
& 85.8{\tiny\color{darkgreen}{$\uparrow$0.5}} 
& 96.3{\tiny\color{darkgreen}{$\uparrow$0.7}} \\

% Row 9: Yes, Yes, No, Yes
\cmark & \cmark & \xmark & \cmark 
& 82.0{\tiny\color{darkred}{$\downarrow$2.2}} 
& 85.5{\tiny\color{darkred}{$\downarrow$2.3}} 
& 93.2{\tiny\color{darkgreen}{$\uparrow$2.2}} 
& 87.2{\tiny\color{darkgreen}{$\uparrow$1.9}} 
& 94.0{\tiny\color{darkred}{$\downarrow$1.6}} \\

% Row 10: Yes, No, Yes, Yes
\cmark & \xmark & \cmark & \cmark 
& 84.0{\tiny\color{darkred}{$\downarrow$0.2}} 
& 90.4{\tiny\color{darkgreen}{$\uparrow$2.6}} 
& 93.4{\tiny\color{darkgreen}{$\uparrow$2.4}} 
& 86.8{\tiny\color{darkgreen}{$\uparrow$1.5}} 
& 95.2{\tiny\color{darkred}{$\downarrow$0.4}} \\

% Row 11: Full OPAL (Yes, Yes, Yes, Yes)
\rowcolor{gray!15}
\cmark & \cmark & \cmark & \cmark 
& 85.6{\tiny\color{darkgreen}{$\uparrow$1.4}} 
& 90.3{\tiny\color{darkgreen}{$\uparrow$2.5}} 
& 92.8{\tiny\color{darkgreen}{$\uparrow$1.8}} 
& 86.9{\tiny\color{darkgreen}{$\uparrow$1.6}} 
& 96.1{\tiny\color{darkgreen}{$\uparrow$0.5}} \\
\bottomrule
\end{tabularx}
\end{table}

        \subsubsection{Architectural Component Analysis.}

\cref{tab:ablation_full} analyzes the contribution of OPAL components by selectively enabling the projection, competitive normalization, pooling, and orthonormal anchoring mechanisms. Since the orthonormal anchors are defined in a \(K=C\cdot m\) space, they require the \(1\times1\) projection to match the channel dimensionality, so configurations with Orth.\ necessarily include Proj.

Two observations are consistent across datasets. First, the orthonormal anchors play a central role: enabling Orth. yields a clear and consistent improvement over the corresponding non-orthogonal variants (+4.07 percentage points in top-1 accuracy on average, across matched settings). Second, pooling-based localization alone is not sufficient. Configurations that rely on GMP without Orth. (rows 3/5/6) incur a substantial average drop (-9.86 percentage points in top-1 accuracy) relative to the baseline, despite producing localized evidence. Consequently, the ablation indicates that OPAL benefits arise from coupling competitive, localized aggregation with the fixed class-structured geometry induced by orthonormal anchoring.

Although certain ablated variants outperform full OPAL on isolated datasets (e.g., Proj+GMP+Orth without CWS in PETS/DOGS, or Proj+CWS+Orth without GMP in CARS/PETS), these configurations compromise key properties that OPAL is designed to provide. In particular, removing GMP breaks the one-maximizer-per-channel mechanism, weakening the direct traceability between each prototype activation and a unique localized image region, which is central to faithful, part-based explanations. Conversely, removing CWS eliminates explicit spatial competition, which empirically increases prototype redundancy and makes it harder to obtain diverse prototypes with distinct semantic roles. Both quantitative and qualitative evidence of the resulting prototype collapse is provided in Sec.~D.1 of the supplementary material. Overall, these findings indicate that strong performance together with reliable prototype diversity and localized interpretability requires combining orthonormal anchoring with spatial competition and max-based aggregation.

\begin{figure}[tb]
          \centering
          \begin{subfigure}{0.3\linewidth}
            \includegraphics[width=\textwidth]{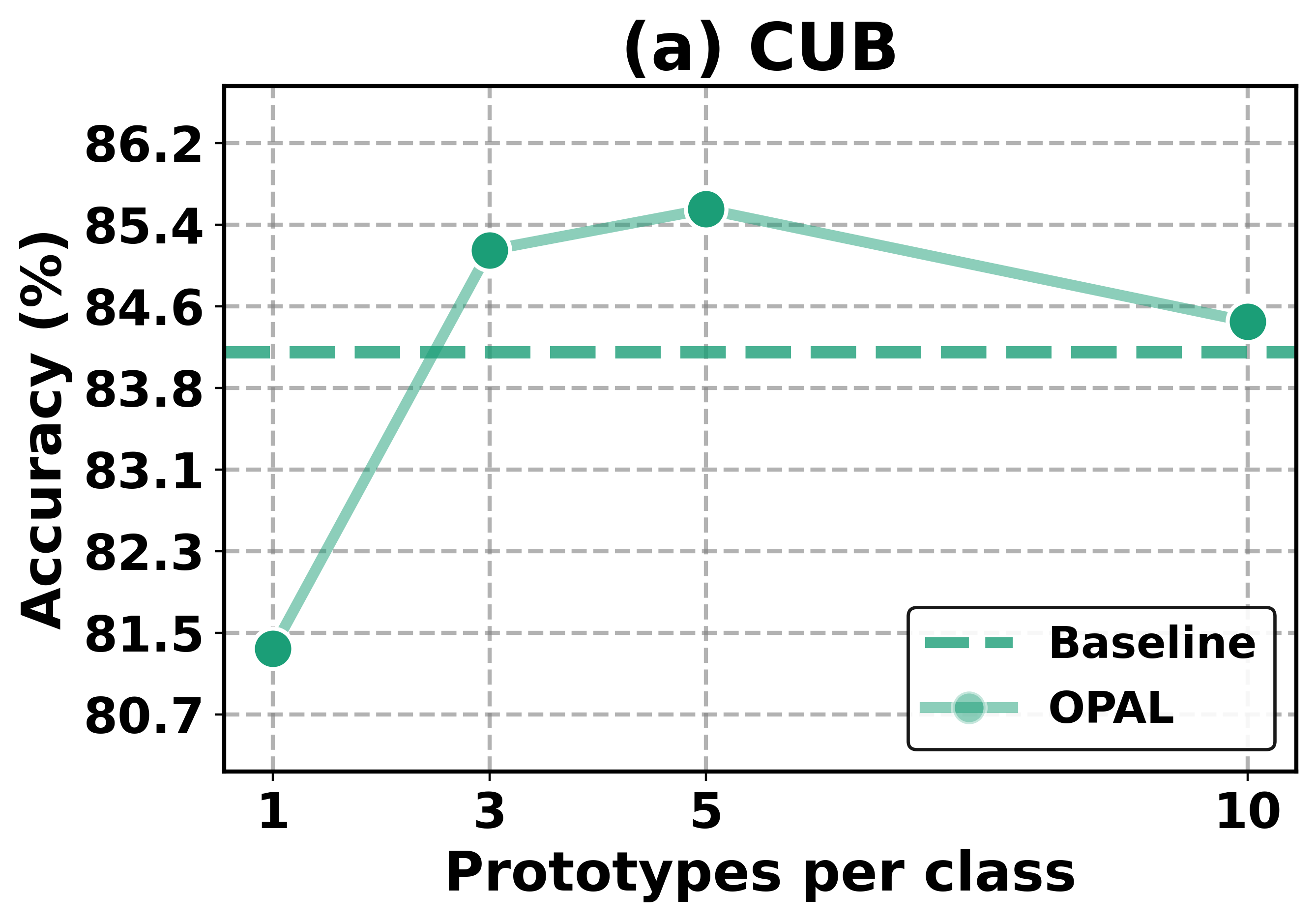}
            \label{subfig:CUB_cpc}
          \end{subfigure}
          \hfill
          \begin{subfigure}{0.3\linewidth}
            \includegraphics[width=\textwidth]{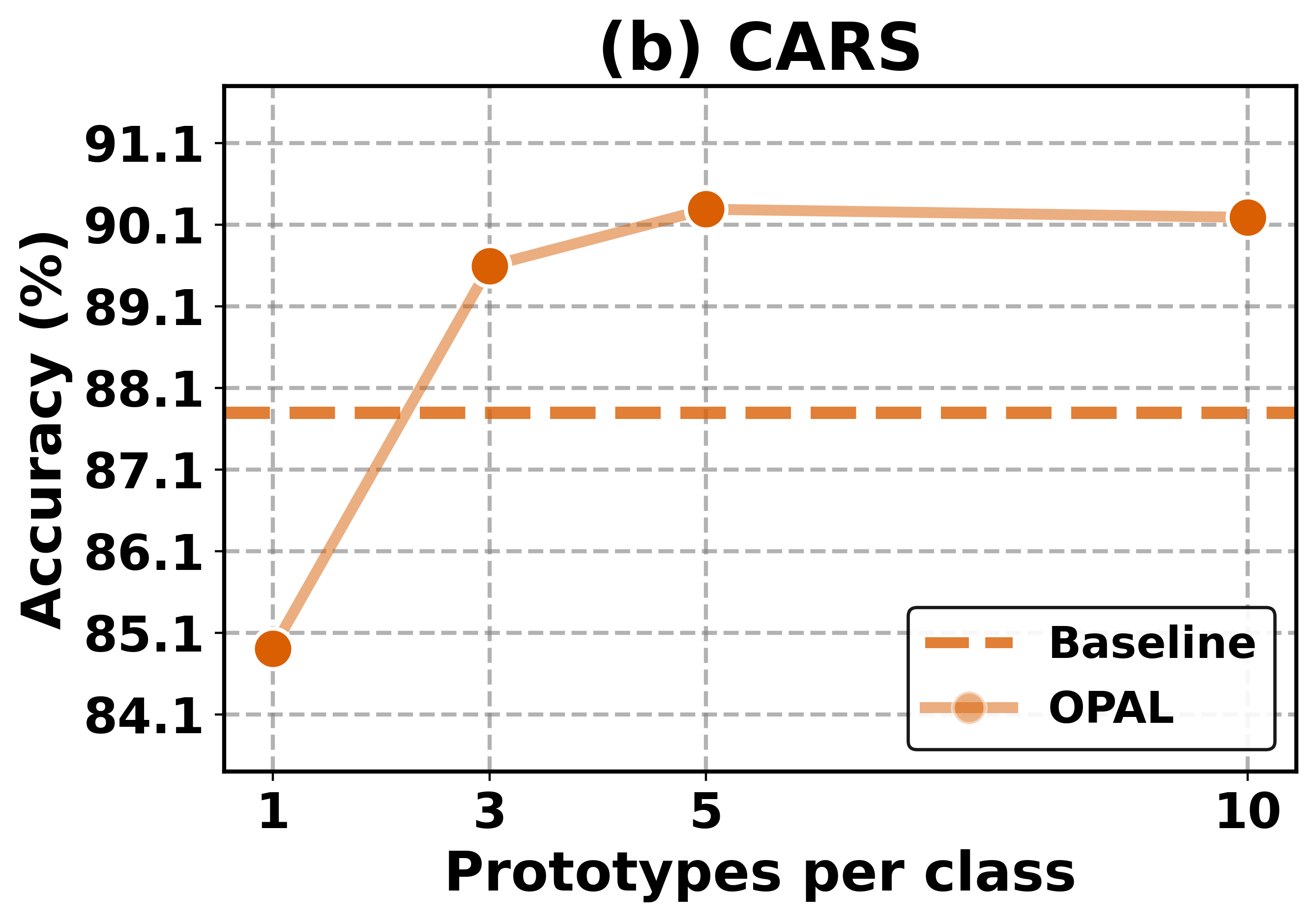}
            \label{subfig:CARS_cpc}
          \end{subfigure}
          \hfill
          \begin{subfigure}{0.3\linewidth}
            \includegraphics[width=\textwidth]{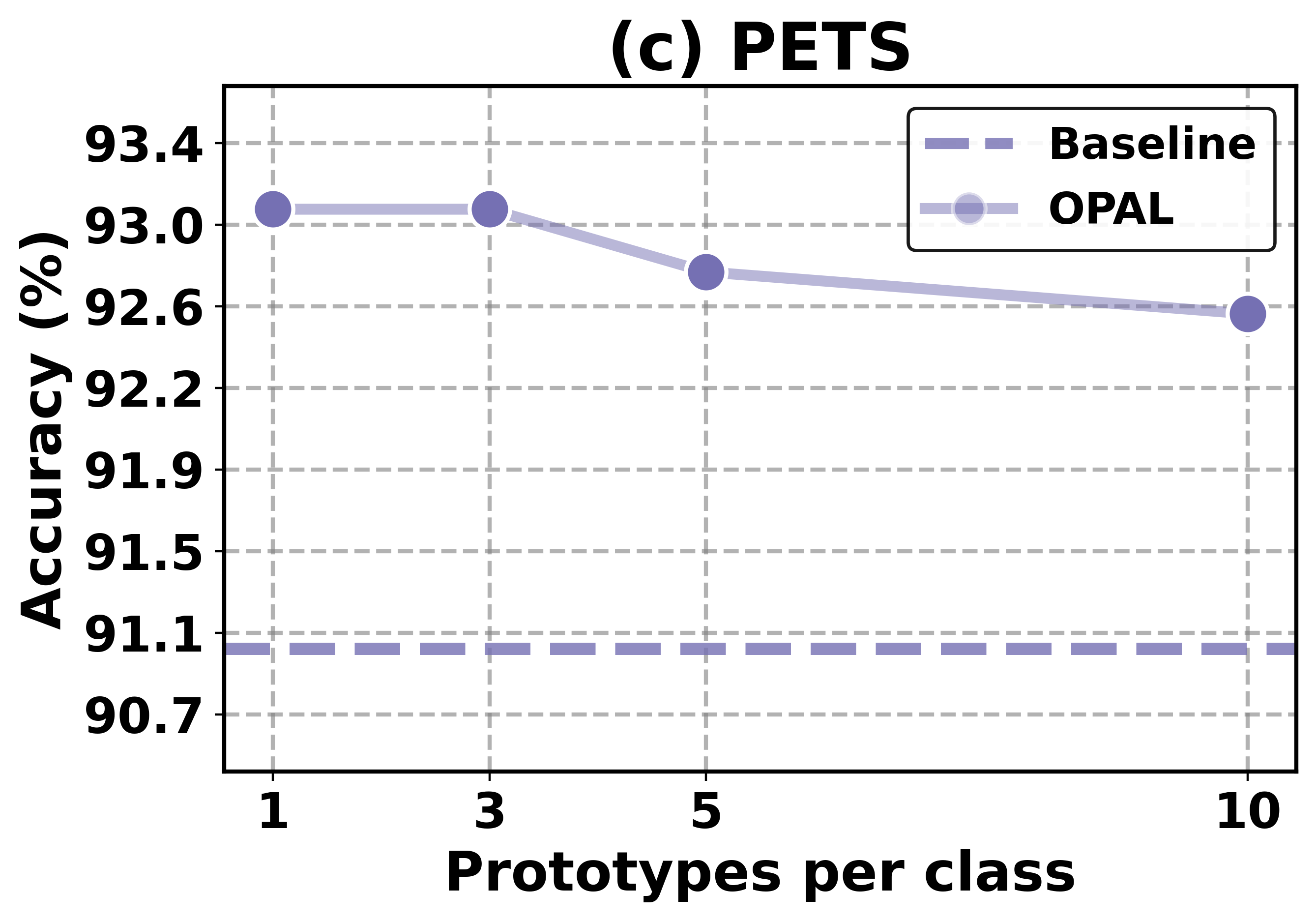}
            \label{subfig:PETS_cpc}
          \end{subfigure}
          \hfill
          \begin{subfigure}{0.3\linewidth}
            \includegraphics[width=\textwidth]{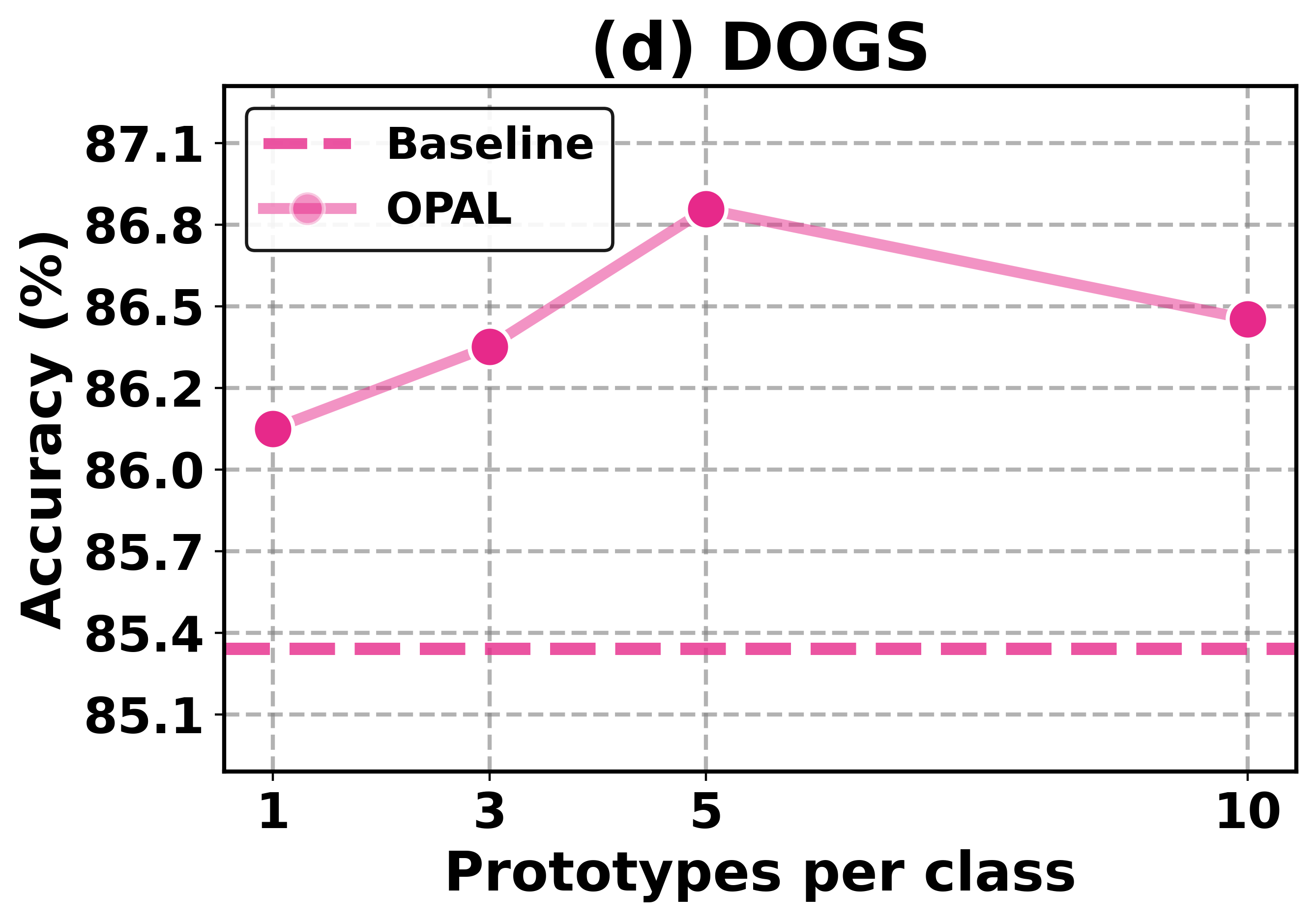}
            \label{subfig:DOGS_cpc}
          \end{subfigure}
          \hfill
          \begin{subfigure}{0.3\linewidth}
            \includegraphics[width=\textwidth]{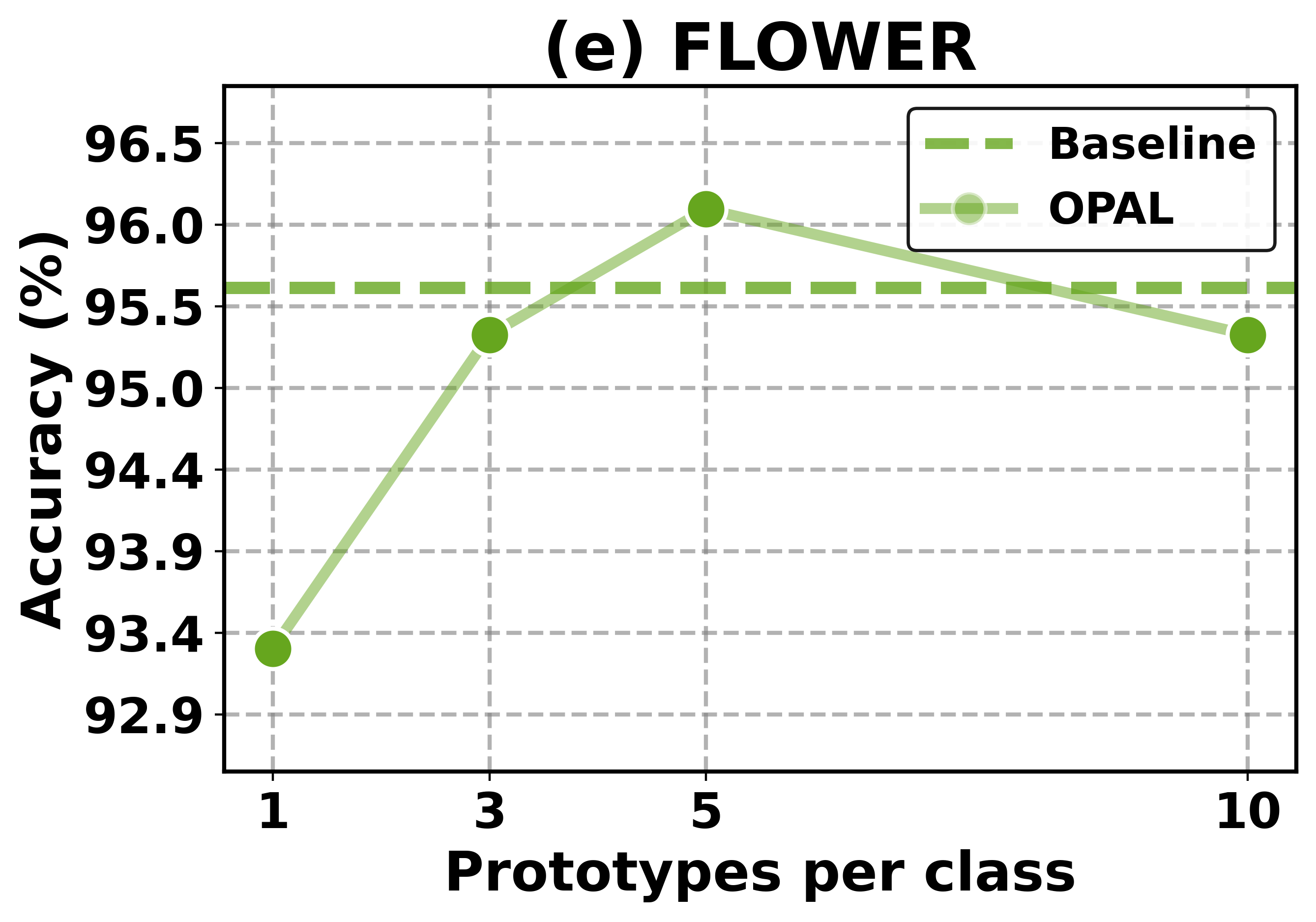}
            \label{subfig:FLOWER_cpc}
          \end{subfigure}
          \hfill
          \begin{subfigure}{0.3\linewidth}
            \includegraphics[width=\textwidth]{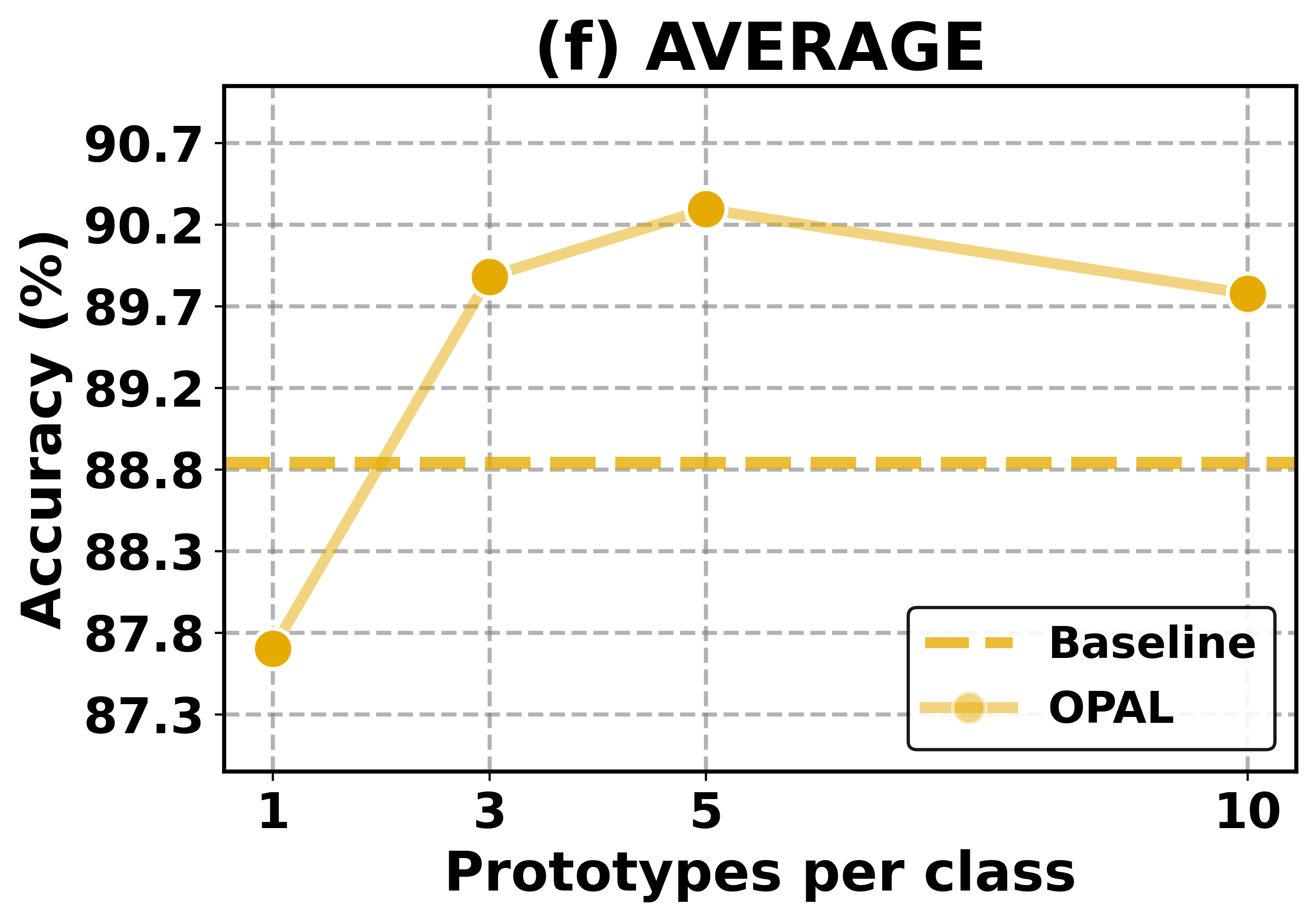}
            \label{subfig:AVG_cpc}
          \end{subfigure}
          \caption{Sensitivity to the number of prototypes per class \(m\). OPAL is evaluated with \(m\in\{1,3,5,10\}\) on (a) CUB, (b) CARS, (c) PETS, (d) DOGS, and (e) FLOWER, with (f) reporting the average across datasets. The dashed line denotes the ConvNeXt-Tiny baseline, and points correspond to OPAL under the same training protocol.}
          \label{fig:prototypes_ablation}
        \end{figure}

        \subsubsection{Sensitivity to the Number of Prototypes}

 We analyze the sensitivity of OPAL to the number of prototypes per class \(m\) by varying \(m\in\{1,3,5,10\}\) while keeping the remaining training protocol fixed. \cref{fig:prototypes_ablation} reports results across datasets and their average. Using a single prototype per class (\(m=1\)) is insufficient and underperforms the ConvNeXt-Tiny baseline by \(-1.04\) percentage points on average. In contrast, selecting \(m\in\{3,5,10\}\) yields consistent gains, with an average improvement of \(+1.25\) percentage points over the baseline when averaging the three configurations.

A clear trend emerges in most benchmarks: allocating too few prototypes limits representational capacity, while an excessive number of prototypes can dilute discriminative evidence, leading to slightly lower accuracy than the best setting. These results indicate that a moderate number of prototypes per class (\(m=5\) in our main experiments) provides a robust operating point, enabling OPAL to deliver interpretable predictions while improving performance over the non-interpretable counterpart. Additional qualitative results for \(m\in\{1,3,5,10\}\) are presented in Sec.~C.2 of the supplementary material.

\section{Conclusion}
\label{sec:experiments}

In this work, we introduced OPAL, an inherently interpretable framework that shifts prototype-based classification from learning latent prototypes to aligning representations within a fixed, class-structured geometry. OPAL anchors each class in a predefined orthonormal subspace and couples this structure with competitive feature aggregation, yielding localized evidence directly tied to the model computation. This design enables a single-stage, end-to-end training pipeline with a standard classification objective, while preserving interpretable, part-based explanations without auxiliary regularizers or multi-stage optimization schedules.

OPAL also has limitations that motivate future research directions. First, the current formulation leverages the inductive bias of convolutional backbones, where locality and hierarchical feature composition naturally ground prototype evidence in the image plane. Extending OPAL to vision transformers will require designing token-level competition and aggregation mechanisms that preserve the benefits of fixed, discriminative latent geometries while remaining competitive and interpretable. Second, OPAL currently uses a fixed number of prototypes per class and does not share prototypes across classes, which can be restrictive when class complexity varies or when attributes recur across categories. Future work will explore adaptive prototype allocation and controlled prototype sharing strategies to better reflect class complexity and cross-class regularities while preserving interpretability and performance guarantees.

\section*{Acknowledgements}
%Please insert your acknowledgments here.
This work has been supported by the Generalitat Valenciana (GVA)
through the project CIPROM/2022/20 (PROMETEO). The contribution of
GJ. Angulo was supported by his AI Chair PR[AI]RIE--PSAI
(ANR-23-IACL-0008, France 2030).

% ---- Bibliografía (única para todo el documento) ----
\bibliographystyle{splncs04}
\bibliography{main}

% ===============================================================
% MATERIAL SUPLEMENTARIO
% ===============================================================

\clearpage

\setcounter{section}{0}
\setcounter{figure}{0}
\setcounter{table}{0}
\setcounter{equation}{0}
\makeatletter
\let\addcontentsline\ORIGaddcontentsline
\makeatother

\setcounter{tocdepth}{2}
\renewcommand{\thesection}{\Alph{section}}

\renewcommand{\thesection}{\Alph{section}}
\renewcommand{\thesubsection}{\thesection.\arabic{subsection}}
\renewcommand{\thesubsubsection}{\thesubsection.\arabic{subsubsection}}
\renewcommand{\thefigure}{\Alph{figure}}
\renewcommand{\thetable}{\Alph{table}}

\begin{center}
  {\Large\bfseries OPAL: Orthonormal Prototype Alignment\\ Learning
   for Interpretable Image Classification\par}
  \vspace{0.5em}
  {\Large Supplementary Material\par}
  \vspace{0.5em}
  {Ilán Carretero\textsuperscript{1}\orcidlink{0009-0003-9474-6515} \quad
   Gustavo Jesús Angulo\textsuperscript{2}\orcidlink{0000-0001-6106-2692}\\[0.3em]
   Rocío del Amor\textsuperscript{1,3}\orcidlink{0000-0002-5342-2093} \quad
   Valery Naranjo\textsuperscript{1,3}\orcidlink{0000-0002-0181-3412}\par}
  \vspace{0.5em}
  {\small
   \textsuperscript{1}\,\textit{CVBLab, HumanTech, Universitat Politècnica de València, Valencia, Spain}\\
   \texttt{\{ilcarjuc,madeam2,vnaranjo\}@upv.es}\\[0.4em]
   \textsuperscript{2}\,\textit{CMA, Mines Paris, PSL University, Sophia Antipolis, France}\\
   \texttt{jesus.angulo\_lopez@minesparis.psl.eu}\\[0.4em]
   \textsuperscript{3}\,\textit{Artikode Intelligence S.L., Valencia, Spain}\par}
\end{center}
\vspace{0.5em}

\supplementarytoc
\vspace{1.5em}

\section{Extended Theoretical Details}
\label{sec:extended-theoretical-details}

\subsection{Formal Justification of the Block-uniform Anchor}
\label{subsec:formal-justification-orthonormal-subspaces}

Below, we formalize the role of the predefined class anchor in promoting distributed evidence within the correct class block, as stated in Section 3.2 of the main paper.

For any embedding \(z \in \mathbb{R}^K\), let \(z_{B_c}\) denote the restriction of \(z\) to the coordinates in block \(B_c\), with all remaining coordinates set to zero. Equivalently, \(z_{B_c}\) is the orthogonal projection of \(z\) onto the class subspace \(\mathcal{S}_c\). We define its in-block energy as:
\[
r_c := \|z_{B_c}\|_2.
\]
Recall that the class anchor is:
\[
a_c = \frac{1}{\sqrt m}\sum_{j=1}^m p_{c,j},
\]
where \(\{p_{c,j}\}_{j=1}^m\) is an orthonormal basis of \(\mathcal{S}_c\). Since OPAL applies channel-wise softmax, global max pooling, and \(L_2\) normalization, the final embedding \(z\) is coordinate-wise non-negative and unit-norm.

\paragraph{\textbf{Proposition.}}
For a fixed class \(c\), suppose that the in-block energy \(r_c\) is held constant. Then the similarity \(\langle z, a_c\rangle\) is maximized if and only if
\[
z_{B_c} = r_c a_c,
\]
that is, when the evidence inside block \(B_c\) is distributed uniformly across its \(m\) slots. In contrast, if the same in-block energy collapses onto a single slot, namely
\[
z_{B_c} = r_c p_{c,j}
\]
for some \(j \in \{1,\dots,m\}\), then
\[
\langle z, a_c\rangle = \frac{r_c}{\sqrt m}.
\]
Hence, for any \(m>1\), the block-uniform anchor assigns strictly higher similarity to a distributed configuration than to a one-slot collapsed one.

\paragraph{\textbf{Proof.}}
Since \(a_c \in \mathcal{S}_c\) and is zero outside block \(B_c\), it follows that:
\[
\langle z, a_c\rangle = \langle z_{B_c}, a_c\rangle.
\]
Now fix \(r_c = \|z_{B_c}\|_2\). By Cauchy--Schwarz,
\[
\langle z_{B_c}, a_c\rangle \le \|z_{B_c}\|_2 \|a_c\|_2 = r_c,
\]
since \(\|a_c\|_2 = 1\). Equality holds if and only if \(z_{B_c}\) is a non-negative scalar multiple of \(a_c\), which under the constraint \(\|z_{B_c}\|_2 = r_c\) is equivalent to \(z_{B_c} = r_c a_c\). Therefore, among all embeddings with fixed in-block energy, the maximum similarity is attained exactly when the evidence is uniformly distributed across the \(m\) slots of block \(B_c\). Now consider the collapsed configuration \(z_{B_c} = r_c p_{c,j}\). Then
\[
\langle z, a_c\rangle
=
\langle r_c p_{c,j}, a_c\rangle
=
r_c \left\langle p_{c,j}, \frac{1}{\sqrt m}\sum_{\ell=1}^m p_{c,\ell}\right\rangle
=
\frac{r_c}{\sqrt m},
\]
by orthonormality of the prototype directions. For \(m=1\), the statement is trivial, since the class block contains a single slot and there is no distinction between distributed and collapsed evidence.\qedsymbol

Since both \(z\) and \(a_c\) are unit-norm, larger similarity \(\langle z,a_c\rangle\) is equivalent to smaller Euclidean distance \(\|z-a_c\|_2\). Therefore, for \(m>1\), OPAL assigns a larger correct-class logit to distributed in-block evidence than to collapsed evidence with the same in-block energy.

\subsection{Canonicity of the Prototype Basis}
\label{supp:basis_choice}

In Sec.~3.2, we fixed the part-prototypes to the canonical directions
$p_{c,j}=e_{(c-1)m+j}$. We show here that this choice is not arbitrary. Under the two
structural requirements imposed by OPAL, the admissible prototype bases reduce to
permutations of the canonical basis, and any such permutation is absorbed by the
learned $1\times1$ projection. The canonical basis is therefore the natural
representative of the only admissible family, rather than one option among many.

The two requirements are the following. First, orthonormality of the prototype family
$\{p_{c,j}\}$, which underpins the geometry of the mutually orthogonal class subspaces
$\{\mathcal{S}_c\}_{c=1}^{C}$ and the maximal angular separation between class anchors.
Second, coordinate-wise non-negativity: the embedding $z\in\mathbb{R}^{K}_{\geq 0}$ is
non-negative by construction, since it is produced by a channel-wise softmax, a global
max pooling, and an $L_2$ normalization (CWS+GMP+L$_2$). The prototype directions, which
the embedding is aligned to, are taken to be non-negative accordingly.

\paragraph{\textbf{Proposition.}} Let $\{p_k\}_{k=1}^{K}\subset\mathbb{R}^{K}_{\geq 0}$ be an
orthonormal family of coordinate-wise non-negative vectors. Then $\{p_k\}_{k=1}^{K}$ is a
permutation of the canonical basis $\{e_k\}_{k=1}^{K}$.

\paragraph{\textbf{Proof.}} Consider any two distinct vectors $p,q$ in the family. Since both are
non-negative, their inner product is a sum of non-negative terms,
\[
\langle p,q\rangle=\sum_{k=1}^{K}p_k\,q_k=0,
\]
so orthogonality forces $p_k\,q_k=0$ for every coordinate $k$, and hence $p$ and $q$ have
disjoint supports. The $K$ vectors thus have pairwise disjoint supports, and since each is unit-norm, every
support is nonempty. We then have $K$ disjoint nonempty subsets within $K$ coordinates,
which fit only if each support is a single coordinate, so the supports partition
$\{1,\dots,K\}$. A unit-norm, non-negative vector supported on one coordinate then equals a
canonical basis vector, and the assignment of coordinates to vectors is a permutation.\qedsymbol

Finally, any such permutation merely relabels the $K$ output coordinates, which amounts
to a row reordering of the $1\times1$ projection producing $G$. As this projection is
learned, the relabeling is absorbed during training and all permutations are equivalent.
We therefore adopt the block-ordered assignment $p_{c,j}=e_{(c-1)m+j}$ as the natural
canonical representative.

\section{Additional Experimental Setup}
\label{sec:additional-experimental-setup}

\subsection{Dataset Details and Preprocessing}
\label{subsec:dataset-details-preprocessing}

We evaluate OPAL on five standard fine-grained benchmarks: CUB-200-2011 \cite{wah2011caltech}, Stanford Cars \cite{krause20133d}, Oxford-IIIT Pets \cite{parkhi2012cats}, Stanford Dogs \cite{khosla2011novel}, and Oxford Flowers-102 \cite{nilsback2008automated}. For all datasets, we follow the official train/test partitions provided by the dataset authors. For datasets without an official validation split, we reserve a stratified 20\% subset of the training data for validation. No custom re-splitting, relabeling, or sample filtering is performed.

All images are resized to a final input resolution of 224×224. Following common practice in prior work \cite{nauta2023pip, wang2024mcpnet, pach2025lucidppn}, all backbones are initialized from ImageNet-pretrained weights, and inputs are normalized using the standard ImageNet mean and standard deviation. During training, we use the TrivialAugment \cite{muller2021trivialaugment} data augmentation scheme, combining geometric and photometric transformations with standard random horizontal flipping, to produce the final 224×224 input. This follows the same general strategy adopted in prior prototype-based methods \cite{nauta2023pip, pach2025lucidppn}.

CUB-200-2011 and Stanford Dogs are cropped to the annotated object bounding boxes prior to training and evaluation, as in prior prototype-based works \cite{nauta2021neural, rymarczyk2022interpretable, nauta2023pip, wang2024mcpnet, pach2025lucidppn}, whereas Stanford Cars, Oxford-IIIT Pets, and Oxford Flowers-102 are used without spatial cropping. Bounding boxes are used only for offline image preprocessing, and no part annotations, segmentation masks, attributes, or other forms of localized supervision are used during training or inference.

\subsection{Implementation Details}
\label{subsec:implementation-details}

All experiments follow the same training protocol across datasets, backbones, and classification heads. Models are fine-tuned end-to-end for 30 epochs with Adam, using a fixed learning rate of \(1\times10^{-4}\) and a batch size of 64. We evaluate ResNet-50/101/152 \cite{he2016deep}, EfficientNet-V2 S/M/L \cite{tan2021efficientnetv2}, and ConvNeXt Tiny/Small/Base \cite{liu2022convnet}, using the official torchvision ImageNet-pretrained weights: \texttt{IMAGENET1K\_V2} for the ResNet models and \texttt{IMAGENET1K\_V1} for the EfficientNet-V2 and ConvNeXt models. The same protocol is used for both the standard linear-head baselines and their OPAL variants.

The final checkpoint is selected on the held-out validation split described in \cref{subsec:dataset-details-preprocessing} according to the lowest validation cross-entropy loss, and final performance is reported as top-1 accuracy on the official test split. Each setting is repeated with three independent random seeds. The main paper reports mean accuracy along with the absolute percentage-point difference relative to the corresponding CNN baseline, whereas the standard deviations across runs are provided in Tables \ref{tab:sota_comparison_supp}, \ref{tab:backbone_scalability_supp}, \ref{tab:ablation_full_supp}, and \ref{tab:sensitivity_cpc}. Results for competing prototype-based methods are obtained from their official repositories and, when available, from the corresponding publications. For reproducibility, all experiments were conducted in Python 3.10 with PyTorch 2.1 and torchvision 0.16 inside the Docker image \texttt{nvcr.io/nvidia/pytorch:23.10-py3}.

\section{Additional Qualitative Interpretability Analysis}
\label{sec:additional-qualitative-interpretability-analysis}

\subsection{Additional Prototype-Based Explanations}
\label{subsec:additional-prototype-based-explanations}

\begin{figure}[H]
  \centering
  \includegraphics[width=\textwidth]{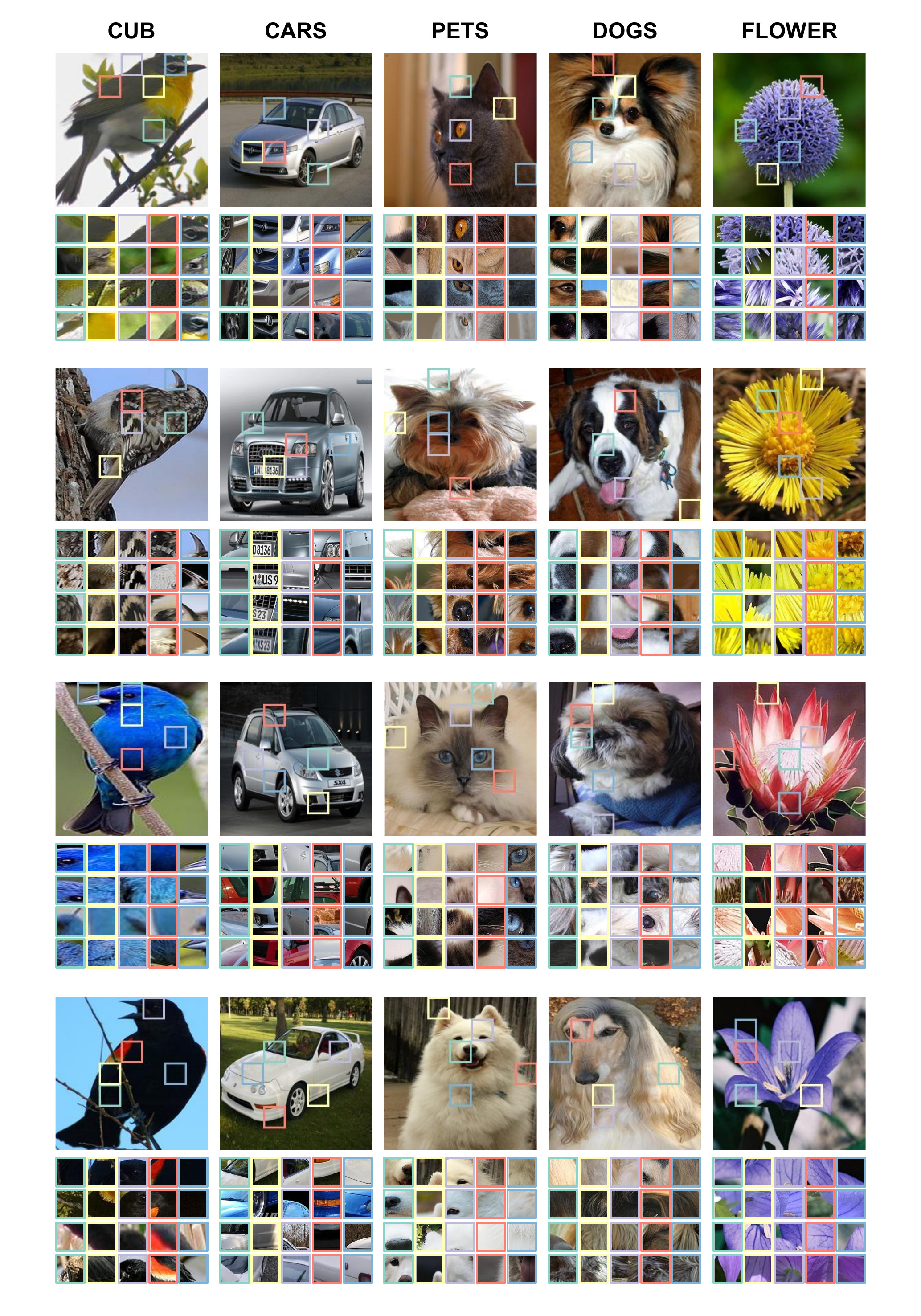}
 \caption{Additional qualitative prototype-based explanations produced by OPAL on five fine-grained benchmarks (\(m=5\)). Colored boxes mark the spatial maximizers selected by each prototype. In the patch grids, the first row is extracted from the displayed test image, while the remaining rows are drawn from other images of the same predicted class.}
\label{fig:qualitative_prototypes_supp}
\end{figure}

\cref{fig:qualitative_prototypes_supp} presents additional qualitative visualizations for \(m=5\). Across all five datasets, the learned prototypes do not collapse and appear to retain clear semantic expressiveness, with different prototype slots consistently attending to distinct and meaningful image regions. Moreover, prototypes associated with different test images from the same class often retrieve highly similar evidence. This suggests that OPAL tends to capture visually consistent and semantically meaningful prototype patterns at the class level.

\subsection{Qualitative Effect of the Number of Prototypes per Class \texorpdfstring{$m$}{m}}
\label{subsec:sensitivity-number-prototypes-per-class}

\begin{figure}[H]
  \centering
  \includegraphics[width=\textwidth]{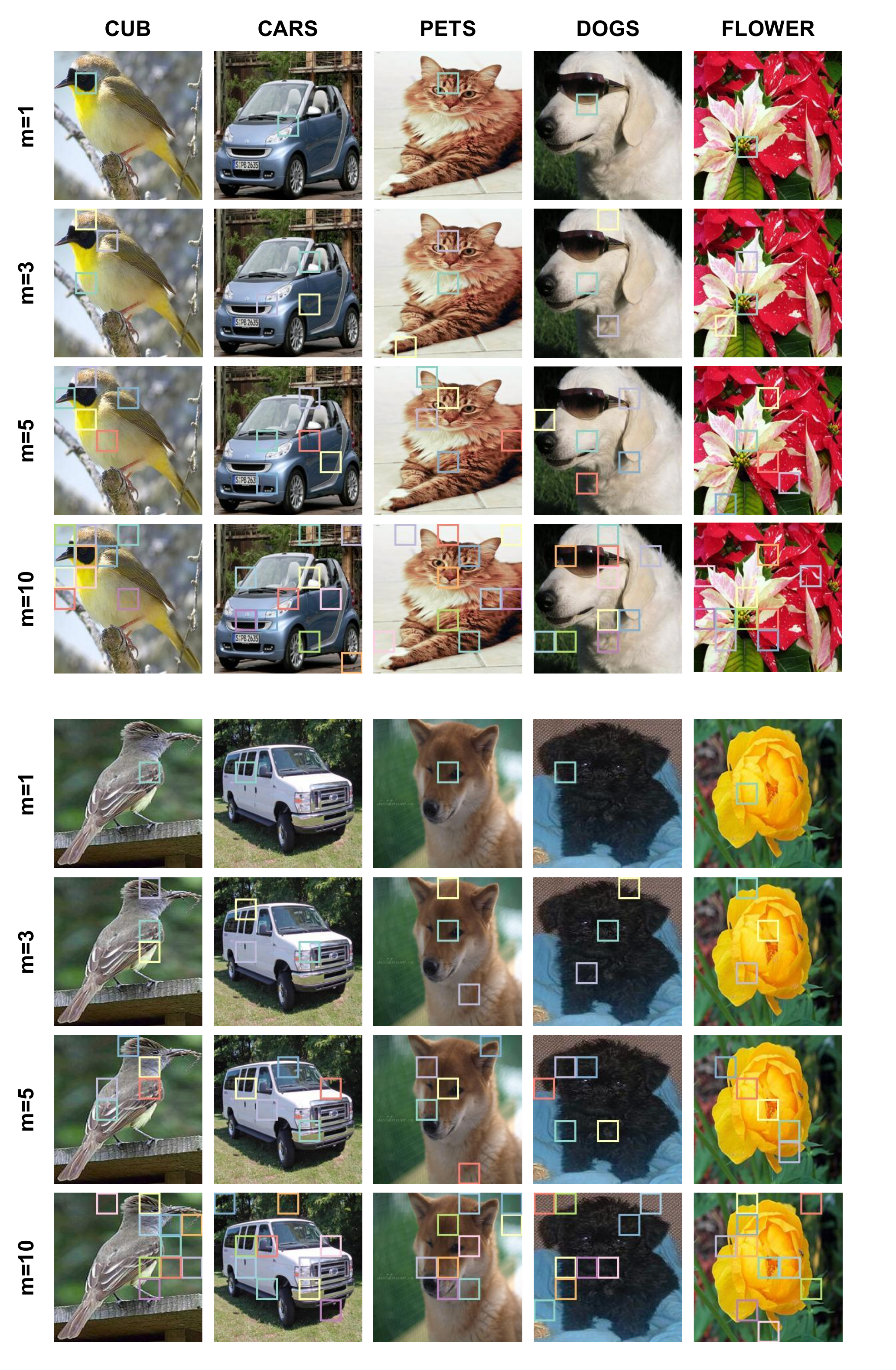}
  \caption{Qualitative prototype-based explanations produced by OPAL on five fine-grained benchmarks for \(m \in \{1,3,5,10\}\). Colored boxes mark the spatial maximizers selected by each prototype.}
\label{fig:qualitative_prototypes_m}
\end{figure}

As illustrated in \cref{fig:qualitative_prototypes_m}, for \(m\in\{1,3,5\}\), the selected prototypes generally attend to semantically meaningful image regions across datasets. When increasing the number of prototypes to \(m=10\), most prototype slots still focus on informative regions, but a small subset begins to attend to areas with limited semantic relevance and, in some cases, even to background regions. This behavior is consistent with the idea that each class contains only a limited number of salient and discriminative regions, meaning that, beyond a certain point, increasing the number of prototypes can lead some slots to capture weakly informative or even null evidence. These qualitative observations are in line with the quantitative trend reported in the main paper (Fig. 4), where moderate values of \(m\) provide the most favorable trade-off between representational capacity and discriminative focus.

\section{Additional Ablation Studies}
\label{sec:additional-ablation-studies}

\subsection{Quantifying Prototype Collapse}
\label{subsec:quantifying-prototype-collapse}

To quantify the prototype redundancy suggested by the ablation experiments in the main paper, we introduce a simple \emph{Spatial Prototype Collapse} (SPC) score. For a test image \(x\) of class \(c\), let \(\mathcal{K}_c\) denote the set of class-specific prototype channels and \((u_k,v_k)=\arg\max A_k\) the spatial maximizer selected by channel \(k\in\mathcal{K}_c\) on the feature map. We define:
\[
\mathrm{SPC}(x)=1-\frac{\left|\{(u_k,v_k):k\in\mathcal{K}_c\}\right|-1}{|\mathcal{K}_c|-1}.
\]
By construction, \(\mathrm{SPC}(x)=0\) means that all class-specific prototypes peak at distinct spatial locations, whereas \(\mathrm{SPC}(x)=1\) indicates complete collapse onto a single feature-map cell. We report the average SPC over the test set, so lower values indicate less collapse and greater spatial diversity.

The results in \cref{tab:spatial_prototype_collapse} reveal a clear effect of \emph{channel-wise softmax} (CWS). When CWS is removed while keeping the projection layer, \emph{global max pooling} (GMP), and the orthonormal basis fixed, spatial collapse becomes severe across all five benchmarks, with SPC values ranging from \(72.1\) to \(93.6\). By contrast, full OPAL reduces SPC to near-zero values on all datasets, indicating that the selected prototypes are typically associated with different spatial positions rather than repeatedly collapsing onto the same one. These results underscore the importance of explicit spatial competition in reducing prototype redundancy.

\begin{figure}[H]
  \centering
  \includegraphics[width=\textwidth]{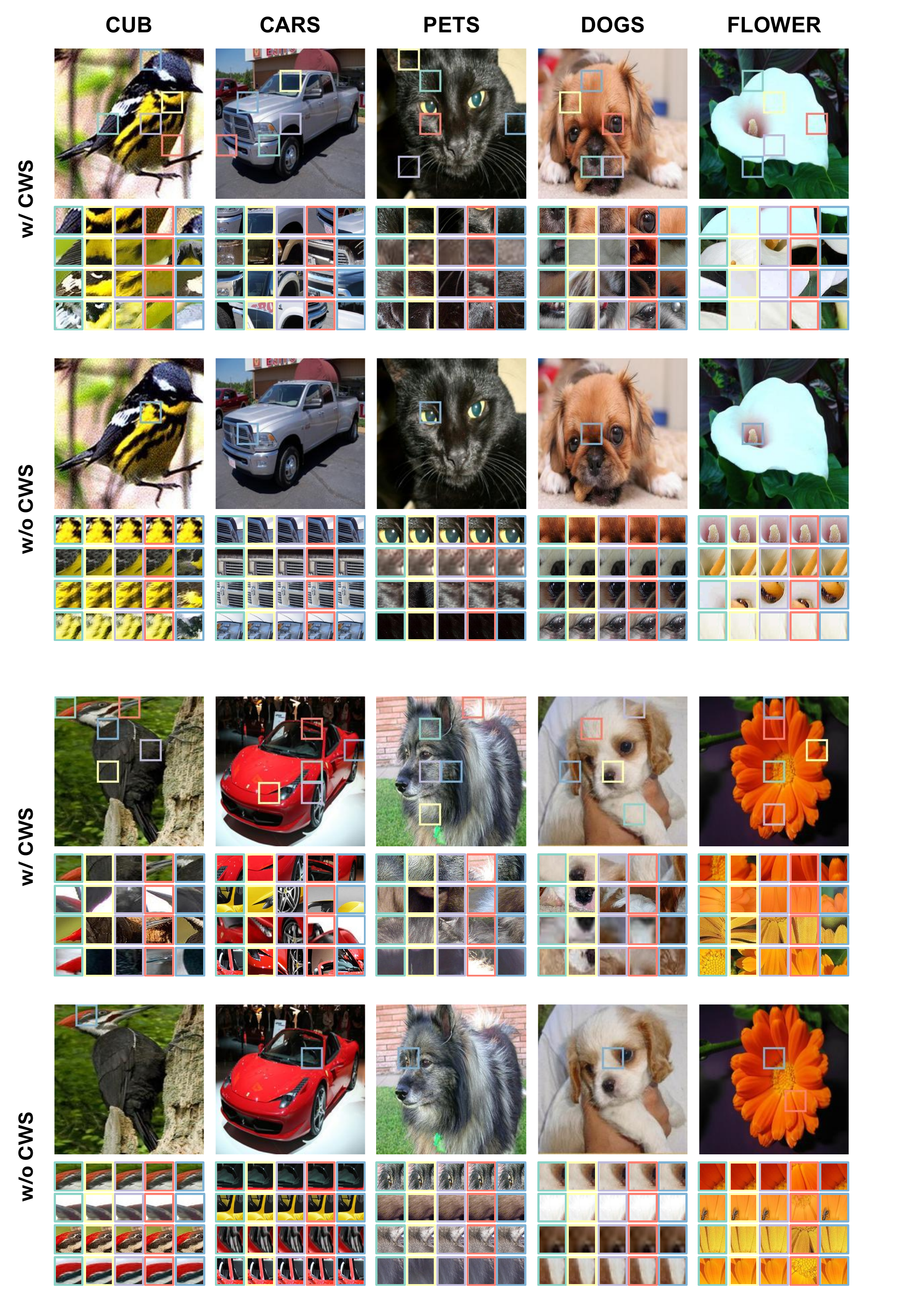}
  \caption{Qualitative prototype-based explanations on five fine-grained benchmarks (\(m=5\)). Columns correspond to datasets. Rows labeled \emph{w/ CWS} show full OPAL, whereas rows labeled \emph{w/o CWS} show the variant without channel-wise softmax. Colored boxes mark the spatial maximizers selected by each prototype.}
\label{fig:qualitative_prototypes_collapse}
\end{figure}

Qualitative examples in \cref{fig:qualitative_prototypes_collapse} reinforce the same conclusion. Without CWS, many prototype slots repeatedly attend to one or two dominant regions, producing highly redundant evidence patches. With CWS, the selected prototypes become substantially more diverse and typically cover distinct, semantically meaningful regions of the object. Overall, the quantitative and qualitative evidence consistently support the combination of the orthonormal basis and CWS, indicating that both components are important for achieving strong performance, preserving inherent interpretability, and preventing prototype collapse.

\begin{table}[tb]
\centering
\caption{Effect of channel-wise softmax on spatial prototype collapse across five fine-grained benchmarks. Columns correspond to the \(1\times1\) convolutional projection (Proj.), global max pooling (GMP), channel-wise softmax (CWS), and orthonormal basis (Orth.). \cmark\ and \xmark\ indicate whether a component is enabled or not. Values are reported as mean \(\pm\) standard deviation, where lower values indicate less spatial prototype collapse.}
\label{tab:spatial_prototype_collapse}
\renewcommand{\arraystretch}{1.2}
\scriptsize

\begin{tabularx}{\textwidth}{@{} cccc CCCCC @{}}
\toprule
\multicolumn{4}{c}{\textbf{Architectural Components}} & \multicolumn{5}{c}{\textbf{Spatial Prototype Collapse (\(\downarrow\))}} \\
\cmidrule(r){1-4} \cmidrule(l){5-9}
\textbf{Proj.} & \textbf{GMP} & \textbf{CWS} & \textbf{Orth.}
& \textbf{CUB} & \textbf{CARS} & \textbf{PETS} & \textbf{DOGS} & \textbf{FLOWER} \\
\midrule

\cmark & \cmark & \xmark & \cmark
& 77.3{\tiny$\pm$21.5}
& 78.7{\tiny$\pm$23.4}
& 93.2{\tiny$\pm$13.9}
& 93.6{\tiny$\pm$12.8}
& 72.1{\tiny$\pm$25.4} \\

\rowcolor{gray!15}
\cmark & \cmark & \cmark & \cmark
& 1.7{\tiny$\pm$8.2}
& 1.4{\tiny$\pm$8.8}
& 1.0{\tiny$\pm$7.2}
& 4.9{\tiny$\pm$15.3}
& 0.1{\tiny$\pm$2.0} \\
\bottomrule
\end{tabularx}
\end{table}

\section{Additional Analyses}
\label{sec:additional-analyses}

\subsection{Alternative Distance Functions}
\label{subsec:alternative-distance-functions}

While the main paper adopts Euclidean distance for class-logit computation, alternative choices can also be considered. Nevertheless, Euclidean distance is particularly well matched to OPAL, since both the embedding \(z\) and the class anchors \(a_c\) are \(L_2\)-normalized, making Euclidean alignment equivalent to cosine-based comparison on the unit hypersphere. For completeness, we therefore evaluate several alternative distances and divergences, namely \(L_1\), \(L_3\), \(L_\infty\) (Chebyshev), Kullback-Leibler divergence, and tropical projective distance \cite{lee2022tropical}. Cosine distance is not included separately because, under the normalized formulation of OPAL, it induces the same ordering as the Euclidean distance.

The results in \cref{tab:distance_sensitivity} show that \(L_1\), \(L_2\), and \(L_3\) consistently provide the strongest performance across datasets. In particular, \(L_1\) and \(L_3\) remain highly competitive, which is consistent with their similarity to \(L_2\) as full-vector discrepancy measures. By contrast, Chebyshev distance performs substantially worse on all benchmarks, while Kullback-Leibler divergence and tropical projective distance are more competitive than Chebyshev but still underperform the \(L_p\) distances with \(p\in\{1,2,3\}\).

This behavior is consistent with the geometry induced by OPAL. The \(L_1\), \(L_2\), and \(L_3\) distances aggregate deviations across the full embedding and therefore remain compatible with the desired alignment of multiple prototype slots within the correct class block. In contrast, \(L_\infty\) depends only on the largest coordinate discrepancy, which is poorly matched to a representation whose decision should reflect joint alignment across several coordinates. Kullback-Leibler divergence and tropical projective distance are likewise less compatible with the normalized orthonormal-anchor formulation, as they emphasize distributional asymmetry or relative coordinate ratios rather than the global geometric alignment captured naturally by the Euclidean distance.

\begin{table}[tb]
\centering
\caption{Sensitivity of OPAL to the choice of distance/divergence function, reported as top-1 accuracy (\%) on five fine-grained benchmarks. \textcolor{darkgreen}{$\uparrow$} and \textcolor{darkred}{$\downarrow$} denote the absolute percentage-point difference with respect to the ConvNeXt-Tiny baseline.}
\label{tab:distance_sensitivity}
\renewcommand{\arraystretch}{1.2}
\scriptsize 

\begin{tabularx}{\textwidth}{@{} l CCCCC @{}}
\toprule
\multirow{2}{*}{\textbf{Method}} & \multicolumn{5}{c}{\textbf{Accuracy ($\uparrow$)}} \\
\cmidrule(l){2-6}
& \textbf{CUB} & \textbf{CARS} & \textbf{PETS} & \textbf{DOGS} & \textbf{FLOWER} \\
\midrule

Baseline (ConvNeXt-Tiny)
& 84.2 
& 87.8 
& 91.0 
& 85.3 
& 95.6 \\

\hdashline\noalign{\vskip 1mm}

OPAL ($L_1$ distance)
& 85.0{\tiny\color{darkgreen}{$\uparrow$0.8}} 
& 90.2{\tiny\color{darkgreen}{$\uparrow$2.4}} 
& 92.6{\tiny\color{darkgreen}{$\uparrow$1.6}} 
& \textbf{88.2}{\tiny\color{darkgreen}{$\uparrow$2.9}} 
& 95.5{\tiny\color{darkred}{$\downarrow$0.1}} \\

OPAL ($L_3$ distance)
& 85.5{\tiny\color{darkgreen}{$\uparrow$1.3}} 
& \textbf{90.4}{\tiny\color{darkgreen}{$\uparrow$2.6}} 
& 92.7{\tiny\color{darkgreen}{$\uparrow$1.7}} 
& 87.7{\tiny\color{darkgreen}{$\uparrow$2.4}} 
& 95.8{\tiny\color{darkgreen}{$\uparrow$0.2}} \\

OPAL ($L_\infty$ distance, Chebyshev)
& 50.5{\tiny\color{darkred}{$\downarrow$33.7}} 
& 46.9{\tiny\color{darkred}{$\downarrow$40.9}} 
& 80.1{\tiny\color{darkred}{$\downarrow$10.9}} 
& 64.3{\tiny\color{darkred}{$\downarrow$21.0}} 
& 73.8{\tiny\color{darkred}{$\downarrow$21.8}} \\

OPAL (Kullback--Leibler divergence)
& 76.7{\tiny\color{darkred}{$\downarrow$7.5}} 
& 80.8{\tiny\color{darkred}{$\downarrow$7.0}} 
& 91.5{\tiny\color{darkgreen}{$\uparrow$0.5}} 
& 86.0{\tiny\color{darkgreen}{$\uparrow$0.7}} 
& 90.6{\tiny\color{darkred}{$\downarrow$5.0}} \\

OPAL (Tropical projective distance)
& 81.5{\tiny\color{darkred}{$\downarrow$2.7}} 
& 79.9{\tiny\color{darkred}{$\downarrow$7.9}} 
& \textbf{92.8}{\tiny\color{darkgreen}{$\uparrow$1.8}} 
& 87.2{\tiny\color{darkgreen}{$\uparrow$1.9}} 
& 90.6{\tiny\color{darkred}{$\downarrow$5.0}} \\

\rowcolor{gray!15}
OPAL ($L_2$ distance, Euclidean)
& \textbf{85.6}{\tiny\color{darkgreen}{$\uparrow$1.4}} 
& 90.3{\tiny\color{darkgreen}{$\uparrow$2.5}} 
& \textbf{92.8}{\tiny\color{darkgreen}{$\uparrow$1.8}} 
& 86.9{\tiny\color{darkgreen}{$\uparrow$1.6}} 
& \textbf{96.1}{\tiny\color{darkgreen}{$\uparrow$0.5}} \\
\bottomrule
\end{tabularx}
\end{table}

\subsection{Computational Efficiency}
\label{subsec:computational-efficiency}

Analyzing computational efficiency is important to assess whether the performance and interpretability gains provided by OPAL come at a meaningful practical cost. \cref{tab:computational_overhead} therefore reports both the complexity of each backbone and the additional dataset-dependent overhead introduced by the standard linear classifier and by OPAL.

Compared with a standard linear head, OPAL incurs a higher computational cost because it replaces the final classifier with a prototype-based head built on a \(1\times1\) projection to \(K=C\cdot m\) channels. In terms of parameters, this overhead is entirely due to the learned projection, since the remaining components are either fixed or parameter-free. In terms of GFLOPs, the \(1\times1\) projection is also expected to dominate, whereas the subsequent operations, including spatial competition and comparisons to fixed orthonormal anchors, remain relatively lightweight. The dataset dependence follows directly from the fact that \(K=C\cdot m\), so the overhead increases with the number of classes and would likewise increase for larger values of \(m\).

More importantly, the dominant computational burden remains the backbone rather than the OPAL head. Across all configurations, the standard linear head accounts for only \(0.04\%\) to \(1.74\%\) of the backbone parameters and less than \(0.015\%\) of its GFLOPs, while OPAL accounts for \(0.20\%\) to \(8.72\%\) of the backbone parameters and \(0.02\%\) to \(2.43\%\) of its GFLOPs. Averaged across all backbones and datasets, the OPAL head represents only \(2.36\%\) of the backbone parameters and \(0.53\%\) of its GFLOPs. Therefore, although OPAL is less efficient than a non-interpretable linear classifier, its additional cost remains small relative to the feature extractor, indicating that the proposed design remains practical from a deployment perspective.

\begin{table}[tb]
\centering
\caption{Computational complexity of the evaluated backbones and the additional dataset-dependent overhead introduced by the standard classification head and OPAL. Backbone columns report the total parameter count and GFLOPs of the feature extractor. For each dataset, we report the additional parameters and GFLOPs of the dataset-specific classification head. Rows named after the backbone correspond to the standard classification head, while shaded rows report the additional overhead introduced by OPAL. Values are expressed in k/M for readability.}
\label{tab:computational_overhead}
\renewcommand{\arraystretch}{1.15}
\scriptsize
\setlength{\tabcolsep}{3.5pt}
\resizebox{\textwidth}{!}{
\begin{tabular}{@{} l cc cc cc cc cc cc @{}}
\toprule
\multirow{3}{*}{\textbf{Backbone}} 
& \multicolumn{2}{c}{\textbf{Backbone}} 
& \multicolumn{10}{c}{\textbf{Additional Overhead}} \\
\cmidrule(lr){2-3}\cmidrule(l){4-13}
& \textbf{Params} & \textbf{GFLOPs} 
& \multicolumn{2}{c}{\textbf{CUB}} 
& \multicolumn{2}{c}{\textbf{CARS}} 
& \multicolumn{2}{c}{\textbf{PETS}} 
& \multicolumn{2}{c}{\textbf{DOGS}} 
& \multicolumn{2}{c}{\textbf{FLOWER}} \\
\cmidrule(lr){4-5}\cmidrule(lr){6-7}\cmidrule(lr){8-9}\cmidrule(lr){10-11}\cmidrule(l){12-13}
&  &  
& \textbf{Params} & \textbf{GFLOPs} 
& \textbf{Params} & \textbf{GFLOPs} 
& \textbf{Params} & \textbf{GFLOPs} 
& \textbf{Params} & \textbf{GFLOPs} 
& \textbf{Params} & \textbf{GFLOPs} \\
\midrule

ResNet-50
& 23.5M & 4.13
& 409.8k & 0.0006
& 401.6k & 0.0006
& 75.8k & 0.0003
& 245.9k & 0.0004
& 209.0k & 0.0004 \\
\rowcolor{gray!15}
\multicolumn{1}{c}{\cellcolor{gray!15}w/ OPAL}
& & 
& 2.049M & 0.1005
& 2.008M & 0.0985
& 379.1k & 0.0186
& 1.229M & 0.0603
& 1.045M & 0.0513 \\

\hdashline\noalign{\vskip 1mm}

ResNet-101
& 42.5M & 7.86
& 409.8k & 0.0006
& 401.6k & 0.0006
& 75.8k & 0.0003
& 245.9k & 0.0004
& 209.0k & 0.0004 \\
\rowcolor{gray!15}
\multicolumn{1}{c}{\cellcolor{gray!15}w/ OPAL}
& & 
& 2.049M & 0.1005
& 2.008M & 0.0985
& 379.1k & 0.0186
& 1.229M & 0.0603
& 1.045M & 0.0513 \\

\hdashline\noalign{\vskip 1mm}

ResNet-152
& 58.1M & 11.60
& 409.8k & 0.0006
& 401.6k & 0.0006
& 75.8k & 0.0003
& 245.9k & 0.0004
& 209.0k & 0.0004 \\
\rowcolor{gray!15}
\multicolumn{1}{c}{\cellcolor{gray!15}w/ OPAL}
& & 
& 2.049M & 0.1005
& 2.008M & 0.0985
& 379.1k & 0.0186
& 1.229M & 0.0603
& 1.045M & 0.0513 \\

\hdashline\noalign{\vskip 1mm}

EfficientNet-V2 S
& 20.2M & 8.90
& 256.2k & 0.0004
& 251.1k & 0.0004
& 47.4k & 0.0002
& 153.7k & 0.0003
& 130.7k & 0.0003 \\
\rowcolor{gray!15}
\multicolumn{1}{c}{\cellcolor{gray!15}w/ OPAL}
& & 
& 1.281M & 0.0629
& 1.255M & 0.0617
& 237.0k & 0.0116
& 768.6k & 0.0377
& 653.3k & 0.0321 \\

\hdashline\noalign{\vskip 1mm}

EfficientNet-V2 M
& 52.9M & 24.54
& 256.2k & 0.0004
& 251.1k & 0.0004
& 47.4k & 0.0002
& 153.7k & 0.0003
& 130.7k & 0.0003 \\
\rowcolor{gray!15}
\multicolumn{1}{c}{\cellcolor{gray!15}w/ OPAL}
& & 
& 1.281M & 0.0629
& 1.255M & 0.0617
& 237.0k & 0.0116
& 768.6k & 0.0377
& 653.3k & 0.0321 \\

\hdashline\noalign{\vskip 1mm}

EfficientNet-V2 L
& 117.2M & 56.12
& 256.2k & 0.0004
& 251.1k & 0.0004
& 47.4k & 0.0002
& 153.7k & 0.0003
& 130.7k & 0.0003 \\
\rowcolor{gray!15}
\multicolumn{1}{c}{\cellcolor{gray!15}w/ OPAL}
& & 
& 1.281M & 0.0629
& 1.255M & 0.0617
& 237.0k & 0.0116
& 768.6k & 0.0377
& 653.3k & 0.0321 \\

\hdashline\noalign{\vskip 1mm}

ConvNeXt-Tiny
& 27.8M & 4.49
& 153.8k & 0.0002
& 150.7k & 0.0002
& 28.5k & 0.0001
& 92.3k & 0.0002
& 78.4k & 0.0002 \\
\rowcolor{gray!15}
\multicolumn{1}{c}{\cellcolor{gray!15}w/ OPAL}
& & 
& 769.0k & 0.0378
& 753.6k & 0.0371
& 142.3k & 0.0070
& 461.4k & 0.0227
& 392.2k & 0.0193 \\

\hdashline\noalign{\vskip 1mm}

ConvNeXt-Small
& 49.5M & 8.74
& 153.8k & 0.0002
& 150.7k & 0.0002
& 28.5k & 0.0001
& 92.3k & 0.0002
& 78.4k & 0.0002 \\
\rowcolor{gray!15}
\multicolumn{1}{c}{\cellcolor{gray!15}w/ OPAL}
& & 
& 769.0k & 0.0378
& 753.6k & 0.0371
& 142.3k & 0.0070
& 461.4k & 0.0227
& 392.2k & 0.0193 \\

\hdashline\noalign{\vskip 1mm}

ConvNeXt-Base
& 87.6M & 15.42
& 205.0k & 0.0003
& 200.9k & 0.0003
& 37.9k & 0.0001
& 123.0k & 0.0002
& 104.6k & 0.0002 \\
\rowcolor{gray!15}
\multicolumn{1}{c}{\cellcolor{gray!15}w/ OPAL}
& & 
& 1.025M & 0.0504
& 1.004M & 0.0494
& 189.6k & 0.0093
& 615.0k & 0.0302
& 522.8k & 0.0257 \\

\bottomrule
\end{tabular}
}
\end{table}

\subsection{Prototype Consistency and Stability on CUB-200-2011}
\label{subsec:prototype-consistency-stability-cub}

While qualitative prototype visualizations are informative, quantitative evaluation is particularly valuable for assessing whether the learned prototypes are semantically meaningful and robust. This type of evaluation requires part-level annotations, which are available only for CUB-200-2011 among the five fine-grained benchmarks considered in this work. Following \cite{huang2023evaluation, pach2025lucidppn}, we therefore report prototype \emph{consistency} and \emph{stability} only on CUB. Consistency measures whether a prototype is repeatedly associated with the same annotated object part across test images from its assigned class, while stability measures whether this association is preserved under input perturbations. In our case, stability is evaluated under the Gaussian random-noise protocol of \cite{huang2023evaluation}. To remain comparable with prior work, we use the previously reported results from \cite{huang2023evaluation, pach2025lucidppn} whenever available, and compute the missing entries under the same evaluation protocol.

The results are summarized in \cref{tab:cub_consistency_stability}. OPAL achieves the second-best score on both metrics, obtaining \(50.5\) in consistency and \(75.9\) in stability. Although this evaluation is limited to a single benchmark and should therefore not be over-generalized, it provides quantitative evidence that OPAL yields reliable prototype explanations under part-level supervision. In particular, OPAL compares favorably with most competing prototype-based models and exhibits a well-balanced quantitative interpretability profile when consistency and stability are considered jointly.

LucidPPN \cite{pach2025lucidppn} attains the highest consistency, which may partly stem from its use of PDiscoNet-generated \cite{van2023pdisconet} part segmentation masks to more explicitly align prototypes with semantic object parts during training \cite{pach2025lucidppn}. While effective, this also introduces dependence on an additional external model for prototype discovery. PIP-Net \cite{nauta2023pip}, in turn, achieves the highest stability, which may be related to its dedicated self-supervised prototype warm-up stage and auxiliary regularization losses. This design, however, makes the training pipeline heavier and introduces additional tuning complexity compared with a single-stage formulation. Overall, OPAL emerges as a particularly strong solution that combines competitive performance and prototype quality with the simplicity of a single-stage end-to-end training pipeline.

\begin{table}[t]
\centering
\caption{Prototype consistency and stability on CUB-200-2011, evaluated following \cite{huang2023evaluation, pach2025lucidppn}. Higher values indicate better semantic alignment and robustness. The best result in each column is highlighted in bold, the second-best is underlined.}
\label{tab:cub_consistency_stability}
\renewcommand{\arraystretch}{1.2}
\scriptsize

\begin{tabularx}{0.7\textwidth}{@{} l CC @{}}
\toprule
\multirow{2}{*}{\textbf{Method}} & \multicolumn{2}{c}{\textbf{CUB}} \\
\cmidrule(l){2-3}
& \textbf{Consistency ($\uparrow$)} & \textbf{Stability ($\uparrow$)} \\
\midrule

ProtoPNet {\tiny\textcolor{gray}{NeurIPS '19}} \cite{chen2019looks}
& 28.3 & 56.7 \\

ProtoTree {\tiny\textcolor{gray}{CVPR '21}} \cite{nauta2021neural}
& 16.4 & 23.2 \\

ProtoPShare {\tiny\textcolor{gray}{KDD '21}} \cite{rymarczyk2021protopshare}
& 9.9 & 58.8 \\

ProtoPool {\tiny\textcolor{gray}{ECCV '22}} \cite{rymarczyk2022interpretable}
& 35.7 & 58.4 \\

PIP-Net {\tiny\textcolor{gray}{CVPR '23}} \cite{nauta2023pip}
& 49.8 & \textbf{89.7} \\

LucidPPN {\tiny\textcolor{gray}{ICLR '25}} \cite{pach2025lucidppn}
& \textbf{71.2} & 66.3 \\

\rowcolor{gray!15}
OPAL \textit{(Ours)}
& \underline{50.5} & \underline{75.9} \\
\bottomrule
\end{tabularx}
\end{table}

The comparison in \cref{tab:cub_consistency_stability} places OPAL among existing methods, but it does not reveal which of its components are responsible for the observed prototype quality. We therefore isolate the two mechanisms that produce localized, non-redundant evidence: CWS, which prevents prototype collapse, and GMP, which anchors each prototype to a single image region. As shown in \cref{tab:cws_gmp_ablation}, removing both (w/o CWS+GMP) yields only a
slight drop in accuracy, yet markedly degrades consistency and stability on CUB-200-2011. Spatial competition and max-based aggregation thus couple the fixed-class geometry to faithful and localized evidence, a quality that accuracy alone does not capture.

\begin{table}[h]
\centering
\caption{Ablation of channel-wise softmax (CWS) and global max pooling (GMP) on accuracy and interpretability metrics on CUB-200-2011. Removing both mechanisms preserves accuracy but markedly degrades prototype consistency and stability. $\uparrow$ denotes the absolute difference with respect to the ablated variant.}
\label{tab:cws_gmp_ablation}
\renewcommand{\arraystretch}{1.2}
\scriptsize
\begin{tabularx}{0.8\textwidth}{@{} l CCC @{}}
\toprule
\multirow{2}{*}{\textbf{Variant}} & \multicolumn{3}{c}{\textbf{CUB}} \\
\cmidrule(l){2-4}
& \textbf{Accuracy ($\uparrow$)} & \textbf{Consistency ($\uparrow$)} & \textbf{Stability ($\uparrow$)} \\
\midrule
w/o CWS + GMP
& 85.0 & 40.3 & 65.1 \\
\rowcolor{gray!15}
OPAL \textit{(Full)}
& \textbf{85.6}{\tiny\color{darkgreen}{$\uparrow$0.6}}
& \textbf{50.5}{\tiny\color{darkgreen}{$\uparrow$10.2}}
& \textbf{75.9}{\tiny\color{darkgreen}{$\uparrow$10.8}} \\
\bottomrule
\end{tabularx}
\end{table}

\subsection{Scalability and Generality Beyond Fine-Grained Benchmarks}
\label{supp:scalability}

We test whether the plug-in behavior of OPAL (main paper, Tab.~3) carries over beyond fine-grained benchmarks, evaluating it on ImageNet-1K (IN-1K) \cite{deng2009imagenet}, Caltech-101 (CT-101) \cite{fei2006onecaltech}, Describable Textures (DTD-47) \cite{cimpoi2014describing}, and Scene UNderstanding (SUN-397) \cite{xiao2010sun}  against its own ConvNeXt-Tiny backbone under an identical protocol (Sec.~\ref{subsec:implementation-details}). \cref{tab:scalability_generality} shows that
OPAL remains competitive with the non-interpretable baseline on IN-1K and CT-101, and consistently improves on DTD-47 and SUN-397. These results suggest that OPAL's benefit is not tied to fine-grained part structure and could extend to broader settings.

\begin{table}[h]
\centering
\caption{Scalability and generality of OPAL beyond fine-grained benchmarks, as top-1 accuracy under the ConvNeXt-Tiny backbone. $\uparrow$/$\downarrow$ denote the percentage-point difference with respect to the baseline.}
\label{tab:scalability_generality}
\renewcommand{\arraystretch}{1.2}
\scriptsize
\begin{tabularx}{0.8\textwidth}{@{} l CCCC @{}}
\toprule
\textbf{Method} & \textbf{IN-1K} & \textbf{CT-101} & \textbf{DTD-47} & \textbf{SUN-397} \\
\midrule
Baseline (ConvNeXt-Tiny)
& 76.7 & 96.9 & 73.2 & 60.3 \\
\rowcolor{gray!15}
OPAL \textit{(Ours)}
& 76.1{\tiny\color{darkred}{$\downarrow$0.6}}
& \textbf{97.2}{\tiny\color{darkgreen}{$\uparrow$0.3}}
& \textbf{76.0}{\tiny\color{darkgreen}{$\uparrow$2.8}}
& \textbf{62.6}{\tiny\color{darkgreen}{$\uparrow$2.3}} \\
\bottomrule
\end{tabularx}
\end{table}

\section{Detailed Quantitative Results}
\label{sec:detailed-quantitative-results}

\subsection{Main-Paper Tables with Mean $\pm$ Standard Deviation Across Three Seeds}
\label{subsec:main-paper-tables-mean-std-three-seeds}

In the main paper, to highlight performance differences in a more direct and visually interpretable manner, we reported not only the mean accuracy across three runs but also the absolute percentage-point difference with respect to the corresponding non-inherently interpretable counterpart. For completeness, Tables \ref{tab:sota_comparison_supp}, \ref{tab:backbone_scalability_supp}, \ref{tab:ablation_full_supp}, and \ref{tab:sensitivity_cpc} present the same quantitative results together with their corresponding standard deviations across three random seeds.

\begin{table}[H]
\centering
\caption{Expanded version of Table 2 in the main paper. Here, the small values reported after each score denote the standard deviation across three random seeds. For methods whose results were taken directly from prior publications, we retain the originally reported values. Consequently, standard deviations are unavailable for some entries. This follows the same reporting practice adopted in prior work \cite{nauta2021neural, rymarczyk2022interpretable, nauta2023pip, wang2024mcpnet, pach2025lucidppn}.}
\label{tab:sota_comparison_supp}
\renewcommand{\arraystretch}{1.2}
\scriptsize 

\begin{tabularx}{\textwidth}{@{} l CCCCC @{}}
\toprule
\multirow{2}{*}{\textbf{Method}} & \multicolumn{5}{c}{\textbf{Accuracy ($\uparrow$)}} \\
\cmidrule(l){2-6}
& \textbf{CUB} & \textbf{CARS} & \textbf{PETS} & \textbf{DOGS} & \textbf{FLOWER} \\
\midrule

% Baseline
Baseline (ConvNext-Tiny)
& 84.2{\tiny$\pm$0.1}  
& 87.8{\tiny$\pm$0.2}  
& 91.0{\tiny$\pm$0.5}  
& 85.3{\tiny$\pm$0.3}  
& 95.6{\tiny$\pm$0.1}  \\

\hdashline\noalign{\vskip 1mm}

% SOTA Models
ProtoPNet {\tiny\textcolor{gray}{NeurIPS '19}} \cite{chen2019looks} 
& 79.2
& 86.1 
& 80.9{\tiny$\pm$0.4} 
& 77.4{\tiny$\pm$0.2} 
& 92.1{\tiny$\pm$0.3} \\

ProtoTree {\tiny\textcolor{gray}{CVPR '21}} \cite{nauta2021neural} 
& 82.2{\tiny$\pm$0.7} 
& 86.6{\tiny$\pm$0.2} 
& 75.0{\tiny$\pm$0.8} 
& 58.9{\tiny$\pm$0.8} 
& 17.5{\tiny$\pm$0.5} \\

ProtoPShare {\tiny\textcolor{gray}{KDD '21}} \cite{rymarczyk2021protopshare} 
& 74.7
& 86.4 
& 81.2{\tiny$\pm$0.4}
& 74.1{\tiny$\pm$0.3} 
& 90.3{\tiny$\pm$0.2} \\

TesNet {\tiny\textcolor{gray}{ICCV '21}} \cite{wang2021interpretable} 
& 84.6{\tiny$\pm$0.3}
& 92.6{\tiny$\pm$0.3} 
& 88.7{\tiny$\pm$0.4}
& 80.7{\tiny$\pm$0.2} 
& 83.4{\tiny$\pm$0.5} \\

ProtoPool {\tiny\textcolor{gray}{ECCV '22}} \cite{rymarczyk2022interpretable} 
& 85.5{\tiny$\pm$0.1} 
& 88.9{\tiny$\pm$0.1} 
& 80.8{\tiny$\pm$0.4} 
& 71.7{\tiny$\pm$0.2} 
& 92.7{\tiny$\pm$0.1} \\

PIP-Net {\tiny\textcolor{gray}{CVPR '23}} \cite{nauta2023pip} 
& 84.3{\tiny$\pm$0.2} 
& 88.2{\tiny$\pm$0.5} 
& 92.0{\tiny$\pm$0.3} 
& 80.8{\tiny$\pm$0.4} 
& 91.8{\tiny$\pm$0.5} \\

MCPNet {\tiny\textcolor{gray}{CVPR '24}} \cite{wang2024mcpnet} 
& 83.5 
& 83.2{\tiny$\pm$0.4} 
& 91.3{\tiny$\pm$0.2} 
& 76.5{\tiny$\pm$0.4} 
& 94.7{\tiny$\pm$0.2} \\

LucidPPN {\tiny\textcolor{gray}{ICLR '25}} \cite{pach2025lucidppn} 
& 81.5{\tiny$\pm$0.4} 
& 91.6{\tiny$\pm$0.2} 
& 92.1{\tiny$\pm$0.3} 
& 79.5{\tiny$\pm$0.4} 
& 95.0{\tiny$\pm$0.3} \\

% Ours
\rowcolor{gray!15}
OPAL \textit{(Ours)} 
& 85.6{\tiny$\pm$0.2} 
& 90.3{\tiny$\pm$0.1} 
& 92.8{\tiny$\pm$0.2} 
& 86.9{\tiny$\pm$0.2} 
& 96.1{\tiny$\pm$0.3} \\
\bottomrule
\end{tabularx}
\end{table}

\begin{table}[H]
\centering
\caption{Expanded version of Table 3 in the main paper. Here, the small values reported after each score denote the standard deviation across three random seeds, whereas the corresponding main-paper table reports the absolute percentage-point difference relative to the corresponding backbone baseline. Shaded rows denote the OPAL variants.}
\label{tab:backbone_scalability_supp}
\renewcommand{\arraystretch}{1.2}
\scriptsize 

\begin{tabularx}{\textwidth}{@{} l CCCCC @{}}
\toprule
\multirow{2}{*}{\textbf{Backbone}} & \multicolumn{5}{c}{\textbf{Accuracy ($\uparrow$)}} \\
\cmidrule(l){2-6}
& \textbf{CUB} & \textbf{CARS} & \textbf{PETS} & \textbf{DOGS} & \textbf{FLOWER} \\
\midrule

% ResNet-50
ResNet-50 
& 72.6{\tiny$\pm$0.2} 
& 71.9{\tiny$\pm$0.2} 
& 89.5{\tiny$\pm$0.2} 
& 81.3{\tiny$\pm$0.3}
& 92.8{\tiny$\pm$0.3}  \\
\rowcolor{gray!15}
\multicolumn{1}{c}{\cellcolor{gray!15}w/ OPAL} 
& 80.8{\tiny$\pm$0.1} 
& 84.3{\tiny$\pm$0.1} 
& 89.8{\tiny$\pm$0.3} 
& 81.0{\tiny$\pm$0.7} 
& 92.7{\tiny$\pm$0.3} \\

\hdashline\noalign{\vskip 1mm}

% ResNet-101
ResNet-101 
& 74.9{\tiny$\pm$0.2}
& 73.8{\tiny$\pm$0.2} 
& 88.5{\tiny$\pm$0.2} 
& 83.5{\tiny$\pm$0.4} 
& 91.6{\tiny$\pm$0.3}  \\
\rowcolor{gray!15}
\multicolumn{1}{c}{\cellcolor{gray!15}w/ OPAL} 
& 81.4{\tiny$\pm$0.2} 
& 83.9{\tiny$\pm$0.1} 
& 91.0{\tiny$\pm$0.3} 
& 83.0{\tiny$\pm$0.7} 
& 93.0{\tiny$\pm$0.3} \\

\hdashline\noalign{\vskip 1mm}

% ResNet-152
ResNet-152 
& 76.0{\tiny$\pm$0.3} 
& 73.7{\tiny$\pm$0.1} 
& 90.3{\tiny$\pm$0.3} 
& 83.1{\tiny$\pm$0.4} 
& 92.4{\tiny$\pm$0.3}  \\
\rowcolor{gray!15}
\multicolumn{1}{c}{\cellcolor{gray!15}w/ OPAL} 
& 80.7{\tiny$\pm$0.1} 
& 85.1{\tiny$\pm$0.2} 
& 90.1{\tiny$\pm$0.3} 
& 84.1{\tiny$\pm$0.6} 
& 93.8{\tiny$\pm$0.2} \\

\hdashline\noalign{\vskip 1mm}

% EfficientNet-V2 S
EfficientNet-V2 S 
& 78.3{\tiny$\pm$0.3} 
& 76.7{\tiny$\pm$0.2} 
& 89.6{\tiny$\pm$0.7} 
& 84.5{\tiny$\pm$0.1} 
& 95.3{\tiny$\pm$0.5}  \\
\rowcolor{gray!15}
\multicolumn{1}{c}{\cellcolor{gray!15}w/ OPAL} 
& 83.5{\tiny$\pm$0.7} 
& 87.2{\tiny$\pm$0.1} 
& 90.4{\tiny$\pm$0.3} 
& 85.8{\tiny$\pm$0.6} 
& 95.3{\tiny$\pm$0.3} \\

\hdashline\noalign{\vskip 1mm}

% EfficientNet-V2 M
EfficientNet-V2 M 
& 79.4{\tiny$\pm$0.3} 
& 80.3{\tiny$\pm$0.2} 
& 89.0{\tiny$\pm$0.2} 
& 83.3{\tiny$\pm$0.1} 
& 93.9{\tiny$\pm$0.9}  \\
\rowcolor{gray!15}
\multicolumn{1}{c}{\cellcolor{gray!15}w/ OPAL} 
& 82.6{\tiny$\pm$0.2} 
& 82.7{\tiny$\pm$0.1} 
& 90.6{\tiny$\pm$0.9} 
& 85.7{\tiny$\pm$0.2} 
& 92.9{\tiny$\pm$0.5} \\

\hdashline\noalign{\vskip 1mm}

% EfficientNet-V2 L
EfficientNet-V2 L 
& 84.1{\tiny$\pm$0.3} 
& 85.9{\tiny$\pm$0.1} 
& 90.0{\tiny$\pm$0.3} 
& 81.4{\tiny$\pm$0.7} 
& 94.9{\tiny$\pm$0.2}  \\
\rowcolor{gray!15}
\multicolumn{1}{c}{\cellcolor{gray!15}w/ OPAL} 
& 86.9{\tiny$\pm$0.1} 
& 89.4{\tiny$\pm$0.2} 
& 90.0{\tiny$\pm$0.1} 
& 83.1{\tiny$\pm$0.8} 
& 96.6{\tiny$\pm$0.4} \\

\hdashline\noalign{\vskip 1mm}

% ConvNeXt-Tiny
ConvNeXt-Tiny 
& 84.2{\tiny$\pm$0.1} 
& 87.8{\tiny$\pm$0.2} 
& 91.0{\tiny$\pm$0.5} 
& 85.3{\tiny$\pm$0.3} 
& 95.6{\tiny$\pm$0.1}  \\
\rowcolor{gray!15}
\multicolumn{1}{c}{\cellcolor{gray!15}w/ OPAL} 
& 85.6{\tiny$\pm$0.2} 
& 90.3{\tiny$\pm$0.1} 
& 92.8{\tiny$\pm$0.2} 
& 86.9{\tiny$\pm$0.2} 
& 96.1{\tiny$\pm$0.3} \\

\hdashline\noalign{\vskip 1mm}

% ConvNeXt-Small
ConvNeXt-Small 
& 81.7{\tiny$\pm$0.2} 
& 83.9{\tiny$\pm$0.2} 
& 91.8{\tiny$\pm$0.1} 
& 86.5{\tiny$\pm$0.4} 
& 95.3{\tiny$\pm$0.1}  \\
\rowcolor{gray!15}
\multicolumn{1}{c}{\cellcolor{gray!15}w/ OPAL} 
& 85.6{\tiny$\pm$0.1} 
& 90.6{\tiny$\pm$0.2} 
& 93.4{\tiny$\pm$0.5} 
& 87.7{\tiny$\pm$0.6} 
& 95.5{\tiny$\pm$0.3} \\

\hdashline\noalign{\vskip 1mm}

% ConvNeXt-Base
ConvNeXt-Base 
& 82.1{\tiny$\pm$0.2} 
& 82.1{\tiny$\pm$0.2} 
& 92.0{\tiny$\pm$0.3} 
& 88.9{\tiny$\pm$0.4} 
& 97.2{\tiny$\pm$0.3}  \\
\rowcolor{gray!15}
\multicolumn{1}{c}{\cellcolor{gray!15}w/ OPAL} 
& 86.3{\tiny$\pm$0.1} 
& 91.4{\tiny$\pm$0.2} 
& 93.9{\tiny$\pm$0.4} 
& 90.2{\tiny$\pm$0.6} 
& 96.6{\tiny$\pm$0.3} \\

\bottomrule
\end{tabularx}
\end{table}

\begin{table}[tb]
\centering
\caption{Expanded version of Table 4 in the main paper. Here, the small values reported after each score denote the standard deviation across three random seeds, whereas the corresponding main-paper table reports the absolute percentage-point difference relative to the baseline. Columns correspond to the \(1\times1\) convolutional projection (Proj.), global max pooling (GMP), channel-wise softmax (CWS), and orthonormal anchors (Orth.). \cmark\ and \xmark\ indicate whether a component is enabled or not. When GMP is disabled, global average pooling is used. The all-crosses configuration (\xmark\xmark\xmark\xmark) corresponds to the convolutional baseline, while the all-checks configuration (\cmark\cmark\cmark\cmark) corresponds to full OPAL.}
\label{tab:ablation_full_supp}
\renewcommand{\arraystretch}{1.2}
\scriptsize

\begin{tabularx}{\textwidth}{@{} cccc CCCCC @{}}
\toprule
\multicolumn{4}{c}{\textbf{Architectural Components}} & \multicolumn{5}{c}{\textbf{Accuracy ($\uparrow$)}} \\
\cmidrule(r){1-4} \cmidrule(l){5-9}
\textbf{Proj.} & \textbf{GMP} & \textbf{CWS} & \textbf{Orth.} 
& \textbf{CUB} & \textbf{CARS} & \textbf{PETS} & \textbf{DOGS} & \textbf{FLOWER} \\
\midrule

% Row 1: Baseline (No, No, No, No)
\xmark & \xmark & \xmark & \xmark 
& 84.2{\tiny$\pm$0.1} 
& 87.8{\tiny$\pm$0.2} 
& 91.0{\tiny$\pm$0.5} 
& 85.3{\tiny$\pm$0.3} 
& 95.6{\tiny$\pm$0.1}  \\

\hdashline\noalign{\vskip 1mm}

% Row 2: Yes, No, No, No
\cmark & \xmark & \xmark & \xmark 
& 82.3{\tiny$\pm$0.4} 
& 86.3{\tiny$\pm$0.3} 
& 92.0{\tiny$\pm$0.1} 
& 86.2{\tiny$\pm$0.3} 
& 96.3{\tiny$\pm$0.1} \\

% Row 3: No, Yes, No, No
\xmark & \cmark & \xmark & \xmark 
& 79.6{\tiny$\pm$0.5} 
& 75.7{\tiny$\pm$0.9} 
& 88.9{\tiny$\pm$0.9} 
& 84.1{\tiny$\pm$0.6} 
& 93.4{\tiny$\pm$0.1} \\

% Row 4: No, No, Yes, No
\xmark & \xmark & \cmark & \xmark 
& 57.4{\tiny$\pm$0.9} 
& 48.9{\tiny$\pm$0.5} 
& 88.1{\tiny$\pm$0.7} 
& 74.6{\tiny$\pm$0.6} 
& 91.5{\tiny$\pm$0.2} \\

% Row 5: Yes, Yes, No, No
\cmark & \cmark & \xmark & \xmark 
& 77.1{\tiny$\pm$0.1} 
& 73.5{\tiny$\pm$0.7} 
& 91.6{\tiny$\pm$0.2} 
& 86.5{\tiny$\pm$0.1} 
& 94.2{\tiny$\pm$0.2} \\

% Row 6: No, Yes, Yes, No
\xmark & \cmark & \cmark & \xmark 
& 54.6{\tiny$\pm$0.9} 
& 31.9{\tiny$\pm$0.2} 
& 86.2{\tiny$\pm$0.1} 
& 80.2{\tiny$\pm$0.2} 
& 86.3{\tiny$\pm$0.4} \\

% Row 7: Yes, No, Yes, No
\cmark & \xmark & \cmark & \xmark 
& 73.7{\tiny$\pm$0.1} 
& 70.5{\tiny$\pm$0.2} 
& 91.1{\tiny$\pm$0.9} 
& 84.7{\tiny$\pm$0.3} 
& 95.0{\tiny$\pm$0.1} \\

% Row 8: Yes, No, No, Yes
\cmark & \xmark & \xmark & \cmark 
& 85.0{\tiny$\pm$0.2} 
& 90.6{\tiny$\pm$0.1} 
& 92.6{\tiny$\pm$0.2} 
& 85.8{\tiny$\pm$0.9} 
& 96.3{\tiny$\pm$0.2} \\

% Row 9: Yes, Yes, No, Yes
\cmark & \cmark & \xmark & \cmark 
& 82.0{\tiny$\pm$0.2} 
& 85.5{\tiny$\pm$0.2} 
& 93.2{\tiny$\pm$0.3} 
& 87.2{\tiny$\pm$0.3} 
& 94.0{\tiny$\pm$0.1} \\

% Row 10: Yes, No, Yes, Yes
\cmark & \xmark & \cmark & \cmark 
& 84.0{\tiny$\pm$0.2} 
& 90.4{\tiny$\pm$0.1} 
& 93.4{\tiny$\pm$0.5} 
& 86.8{\tiny$\pm$0.3} 
& 95.2{\tiny$\pm$0.1} \\

% Row 11: Full OPAL (Yes, Yes, Yes, Yes)
\rowcolor{gray!15}
\cmark & \cmark & \cmark & \cmark 
& 85.6{\tiny$\pm$0.2} 
& 90.3{\tiny$\pm$0.1} 
& 92.8{\tiny$\pm$0.2} 
& 86.9{\tiny$\pm$0.2} 
& 96.1{\tiny$\pm$0.3} \\
\bottomrule
\end{tabularx}
\end{table}

\begin{table}[tb]
\centering
\caption{Tabular version of Fig. 4. While the sensitivity analysis is presented graphically in the main paper, we provide the corresponding numerical values here for completeness and ease of reference. All values are reported as mean \(\pm\) standard deviation across three random seeds.}
\label{tab:sensitivity_cpc}
\renewcommand{\arraystretch}{1.2}
\scriptsize 

\begin{tabularx}{\textwidth}{@{} l CCCCC @{}}
\toprule
\multirow{2}{*}{\textbf{Method}} & \multicolumn{5}{c}{\textbf{Accuracy ($\uparrow$)}} \\
\cmidrule(l){2-6}
& \textbf{CUB} & \textbf{CARS} & \textbf{PETS} & \textbf{DOGS} & \textbf{FLOWER} \\
\midrule

Baseline (ConvNext-Tiny)
& 84.2{\tiny$\pm$0.1}  
& 87.8{\tiny$\pm$0.2}  
& 91.0{\tiny$\pm$0.5}  
& 85.3{\tiny$\pm$0.3}  
& 95.6{\tiny$\pm$0.1}  \\

\hdashline\noalign{\vskip 1mm}

w/ OPAL $m$=1
& 81.3{\tiny$\pm$0.1} 
& 84.9{\tiny$\pm$0.7} 
& 93.1{\tiny$\pm$0.6} 
& 86.1{\tiny$\pm$0.3} 
& 93.3{\tiny$\pm$0.1} \\

w/ OPAL $m$=3
& 85.2{\tiny$\pm$0.1} 
& 89.6{\tiny$\pm$0.4} 
& 93.1{\tiny$\pm$0.5} 
& 86.4{\tiny$\pm$0.4} 
& 95.3{\tiny$\pm$0.1} \\

\rowcolor{gray!15}
w/ OPAL $m$=5
& 85.6{\tiny$\pm$0.2} 
& 90.3{\tiny$\pm$0.1} 
& 92.8{\tiny$\pm$0.2} 
& 86.9{\tiny$\pm$0.2} 
& 96.1{\tiny$\pm$0.3} \\

w/ OPAL $m$=10
& 84.5{\tiny$\pm$0.3} 
& 90.2{\tiny$\pm$0.2} 
& 92.6{\tiny$\pm$0.2} 
& 86.5{\tiny$\pm$0.4} 
& 95.3{\tiny$\pm$0.1} \\

\bottomrule
\end{tabularx}
\end{table}

% ---- Bibliography ----
%
% BibTeX users should specify bibliography style 'splncs04'.
% References will then be sorted and formatted in the correct style.
%
\end{document}